\documentclass[runningheads,a4paper]{llncs}

\usepackage{url}
  
\makeatletter
\newif\if@restonecol
\makeatother

\usepackage{times}
\usepackage{helvet}
\usepackage{courier}
\usepackage[figure, noend]{algorithm2e}  
\usepackage{stmaryrd}
\usepackage{latexsym} 
\usepackage{amstext}
\usepackage{amsmath}
\usepackage{amsfonts}
\usepackage{amssymb}
\usepackage{xspace}
\usepackage{soul}\setuldepth{x} %% for nice underlying \ul{xxx}
\usepackage{multirow}
\usepackage{subfigure}
\usepackage[sectionbib,sort&compress,numbers,square]{natbib}
\newcommand{\citea}[1]{\citeauthor{#1}~\cite{#1}}

\usepackage{tikz}
\usetikzlibrary{petri,arrows,snakes,backgrounds,matrix,automata,shapes,shadows,patterns,fit,calc,patterns}

\def\newblock{\hskip .11em plus.33em minus.07em}

\newcommand{\comment}[1]{}

\newcommand{\union}{\cup}

\newcommand{\tup}[1]{\langle #1\rangle}            % tuple

\newcommand{\set}[1]{\{ #1 \}}
\newcommand{\A}{\mathcal{A}} 
\newcommand{\C}{\mathcal{C}} \newcommand{\D}{\mathcal{D}}
 \newcommand{\F}{\mathcal{F}}
\newcommand{\G}{\mathcal{G}}

\newcommand{\M}{\mathcal{M}} 
 \renewcommand{\P}{\mathcal{P}}

\newcommand{\U}{\mathcal{U}} \newcommand{\V}{\mathcal{V}}

\newcommand{\commentout}[1]{}

\newcommand{\mathname}[1]{\ensuremath{\operatorname{\textit{#1}}}}
\newcommand{\mathnamealt}[1]{\ensuremath{\operatorname{\textsf{\small #1}}}}

\newcommand{\mapCap}{\ensuremath{\Theta}}
\newcommand{\agents}{\A}
\newcommand{\agt}{{\mathname{agt}}}

\renewcommand{\path}{\ensuremath{\lambda}}

\newcommand{\mapFunc}{\ensuremath{\V}}
\newcommand{\actions}{\mathname{Act}}
\newcommand{\out}{\mathname{\textsf{\small out}}}
\newcommand{\act}{\mathname{a}}
\newcommand{\atld}{\ensuremath{d}}
\newcommand{\transFunc}{\ensuremath{\sigma}}
\newcommand{\goalFunc}{g}

\newcommand{\moveVector}{\ensuremath{\vec{\act}}}
\newcommand{\plan}[1]{\ensuremath{\phi_{#1}[\alpha]_{#1}\psi_{#1}}}
\newcommand{\planf}[3]{\ensuremath{#1[#2]#3}}

\newcommand{\coalition}[1]{\ensuremath{\langle\!\langle #1 \rangle\!\rangle}}
\newcommand{\coalitione}[2]{\coalition{#1}_{#2}}

\newcommand{\nop}{\ensuremath{\textsc{noOp}}}

\newcommand{\BDIOURS}{\mbox{BDI-ATLES}\xspace}
\newcommand{\STRATEGYOURS}{\ensuremath{\langle\omega,\varrho \rangle}}
\newcommand{\outf}{\mathname{\textsf{\small out}}}
\newcommand{\exec}{\mathname{Exec}}

\usepackage{marginnote}
\edef\marginnotetextwidth{2\the\textwidth}

\usepackage{wasysym}
\newcommand{\mnext}{\!\ensuremath{\bigcirc}\!}
\newcommand{\always}{\Box}

\newcommand{\until}{\ensuremath{\U}}

\newcommand{\propername}[1]{\text{\textsc{#1}}\xspace}

\newcommand{\JASON}{\propername{Jason}}

\newcommand{\JACK}{\propername{Jack}}

\newcommand{\DAPL}{\propername{2apl}}
\newcommand{\GOALBDI}{\propername{Goal}}

\newcommand{\JADEX}{\propername{Jadex}}

\newcommand{\under}[1]{\mbox{\underline{\it\smash{#1}\vphantom{\lower.05ex\hbox{
x}}}}}

\newcommand{\defterm}[1]{\emph{\ul{#1}}}
\newcommand{\Navigation}{\mathname{Nav}\xspace}
\renewcommand{\Navigation}{\mathnamealt{Navigate}\xspace}
\newcommand{\Pick}{\mathnamealt{Collect}\xspace}
\newcommand{\Drop}{\mathnamealt{Deposit}\xspace}

\newcommand{\card}[1]{|#1|}

\newcommand{\tr}{\mathname{GetTrace}}

\newcommand{\agB}{\Ag_B}
\newcommand{\agC}{\Ag_C}

\newcommand{\enA}{\En_A}
\newcommand{\enB}{\En_B}
\newcommand{\enC}{\En_C}

\newcommand{\gAg}{G_{\Ag}}
\newcommand{\gEn}{G_{\En}}
\newcommand{\gA}{\ensuremath{G_A}}
\newcommand{\gB}{\ensuremath{G_B}}
\newcommand{\gC}{\ensuremath{G_C}}
\newcommand{\gD}{\ensuremath{G_B}}
\newcommand{\WIN}{\operatorname{\emph{WIN}}}
\newcommand{\WAg}{\psi_{\WIN}}

\newcommand{\true}{\propername{True}}
\newcommand{\false}{\propername{False}}
\newcommand{\separation}[1]{\addtolength{\itemsep}{#1}}

\usepackage[normalem]{ulem}

\begin{document}

\mainmatter  % start of an individual contribution

\title{Reasoning about Agent Programs using ATL-like Logics}

%\titlerunning{Lecture Notes in Computer Science: Authors' Instructions}

\author{Nitin Yadav and Sebastian Sardina 
\thanks{We would like to thank Lawrence Cavedon for earlier discussions on the topic of this paper. We acknowledge the support of the Australian Research Council under grant DP120100332.}
}

\institute{RMIT University, Melbourne, Australia.}

 
\maketitle

\begin{abstract} 
We propose a variant of Alternating-time Temporal Logic (ATL) grounded in the agents' operational know-how, as defined by their libraries of abstract plans.
Inspired by ATLES, a variant itself of ATL, it is possible in our logic to explicitly refer to ``rational" strategies for agents developed under the Belief-Desire-Intention agent programming paradigm. 
This allows us to express and verify properties of BDI systems using ATL-type logical frameworks.
\end{abstract}

%\vspace{-.75cm}
\section{Introduction}\label{sec:intro}

The formal verification of agent-oriented programs requires logic frameworks capable of representing and reasoning about agents' abilities and capabilities, and the goals they can feasibly achieve. 
In particular, we are interested here in programs written in the family of Belief-Desire-Intention (BDI) agent programming systems~\cite{Bratman:CI88,Rao:KR92,Bordinietal:Informatica2006-SurveyLangMAS}, a popular paradigm for building multi-agent systems.
Traditional BDI logics based on CTL (e.g., \cite{Rao:KR91}) are generally too weak for representing ability; their success has primarily been in defining ``rationality postulates,'' i.e., constraints on rational behaviour. Further, such logics do not encode agents' capabilities (as represented by their plan libraries) and thereby leave a sizable gap between agent programs and their formal verification.

Recent work (e.g., \cite{Alechina:AAAI07,Alechina:KR08,Dastani:AAMAS10}) has better bridged the gap between formal logic and practical programming by providing an axiomatisation of a class of models that is designed to closely model a programming framework. However, this is done by restricting the logic's models to those that satisfy the transition relations of agents' plans, as defined by the semantics of the programming language itself. In such a framework, it is not possible to reason about the agent's know-how and what the agent could achieve \emph{if it had} specific capabilities.  It is also not possible to reason about coalition of agents.

% Our aim in this paper is to define a framework that will allow us to speculate about an agent's capabilities and what it may achieve with such capabilities, within a BDI multi-agent context. 
% %%
% In a nutshell, BDI agent systems enable abstract plans written by programmers to be combined and used in real-time, under the principles of practical reasoning~\cite{Bratman:87}.
% %%
% Further, we extend this to coalitions of BDI-type agents---i.e., for a given set of agents, each with a certain specified set of capabilities, what could that coalition achieve (regardless of what other agents outside the coalition may do).

Our aim thus is to define a framework, together with model checking techniques, that will allow us to speculate about a group of agents' capabilities and what they can achieve with such capabilities under the BDI paradigm, which enables abstract plans written by programmers to be combined and used in real-time under the principles of %practical reasoning~\cite{Bratman:CI88}.

This requires the ability to represent capabilities directly in our logic. To that end, we adapt ATLES, a version of ATL (Alternating-time Temporal Logic)~\cite{AluraHK:JACM02-ATL} with  Explicit Strategies~\cite{Walther:TARK07-ATLES}, to our purpose. %%
ATL is a logic for reasoning about the ability of agent coalitions in \emph{multi-player game structures}.
% and has recently proved to be a useful framework for the formal verification of such systems. 
%%
This is achieved by reasoning about strategies (and their success) employed by teams of agents: $\coalition{A}\varphi$ expresses that the coalition team of agents $A$ has a joint strategy for guaranteeing that the temporal property $\varphi$ holds. 
%%
%% Further, the model-checking problem in ATL is, as for CTL, decidable
%% in polynomial time with respect to the size of the ATL structure.
%%
\citea{Walther:TARK07-ATLES}, standard ATL does not allow agents' strategies to be explicitly represented in the syntax of the logic. They thus rectified this shortcoming by defining ATLES, which extends ATL by allowing strategy terms in the language: $\coalitione{A}{\rho}\varphi$ holds if coalition $A$ has a joint strategy for ensuring $\varphi$, when some agents are committed to \emph{specific} strategies as specified by so-called commitment function $\rho$. 
%%
% The authors provided a sound and complete axiomatization of ATLES and show that the complexity of the satisfiability problem is no worse than for ATL.
% 
% To fit our aim of a verification framework for BDI multi-agent systems, we adapt ATLES so that the strategy terms of the language are tied directly to the plans available to each agent via the standard concept of practical reasoning embodied by BDI paradigm~\cite{Bratman:87,Rao:KR92,Bordinietal:Informatica2006-SurveyLangMAS}: \emph{the only strategies that can be employed by a BDI agent are those that ensue by the rational execution of its pre-defined plans, given its goals and beliefs}. 
% %%

In this paper, we go further and develop a framework---called \BDIOURS---in which the strategy terms are tied directly to the plans available to agents under the notion of practical reasoning embodied by the BDI paradigm~\cite{Bratman:CI88,Rao:KR92}: \emph{the only strategies that can be employed by a BDI agent are those that ensue by the (rational) execution of its predefined plans, given its goals and beliefs}. 
The key construct $\coalitione{A}{\omega,\varrho}\varphi$ in the new framework states that coalition $A$ has a joint strategy for ensuring $\varphi$, \emph{under the assumptions that some agents in the system are BDI-style agents} with capabilities and goals as specified by assignments $\omega$ and $\varrho$, respectively.
For instance, in the Gold Mining domain from the International Agent Contest,\footnote{\small http://www.multiagentcontest.org/} 
one may want to verify if two miner agents programmed in a BDI language can successfully collect gold pieces when equipped with navigation and communication capabilities and want to win the game, while the opponent agents can perform any physically legal action. 
More interesting, a formula like $\coalitione{A}{\emptyset,\emptyset}\varphi \supset \coalitione{A}{\omega,\varrho}\varphi$ can be used to check whether coalition $A$ has enough know-how and motivations to carry out a task $\varphi$ that is indeed physically feasible for the coalition.

We observe that the notion of ``rationality'' used in this work is that found in the literature on BDI and agent programming, rather than that common in game-theory (generally captured via \emph{solution concepts}). 
As such, rationality shall refer from now on to reasonable constraints on how the various mental modalities---e.g., beliefs, intention, goals---may interact. 
In particular, we focus on the constraint that agents select actions from their know-how in order to achieve their goals in the context of their beliefs.

Finally, we stress that this work aims to contribute to the agent-oriented programing community more than to the (ATL) verification one.
Indeed, our aim is to motivate the former to adopt well-established techniques in game-theory for the effective verification of their ``reactive'' style agent programs.

\newcommand{\Ag}{\operatorname{\textsf{Ag}}}
\newcommand{\En}{\operatorname{\textsf{En}}}

\newcommand{\actionfont}[1]{\text{\textsc{#1}}}
\newcommand{\aDrop}{\actionfont{drop}\xspace}
\newcommand{\aLeft}{\actionfont{left}\xspace}
\newcommand{\aRight}{\actionfont{right}\xspace}
\newcommand{\aPick}{\actionfont{pick}\xspace}

\section{Preliminaries}\label{sec:preliminaries}

\subsection{ATL/ATLES Logics of Coalitions}\label{sec:atl}

Alternating-time Temporal Logic (ATL)~\cite{AluraHK:JACM02-ATL} is a logic for reasoning about the ability of agent coalitions in \emph{multi-agent game structures}.
% %
ATL formulae are built by combining propositional formulas, the usual temporal operators---namely, $\bigcirc$ (``in the next state''), $\Box$ (``always''), $\Diamond$ (``eventually''), and $\U$ (``strict until'')---and a \emph{coalition} \emph{path quantifier} $\coalition{A}$ taking a set of agents $A$ as parameter. As in CTL, which ATL extends, temporal operators and path quantifiers are required to alternate.
Intuitively, an ATL formula $\coalition{A}\phi$, where $A$ is a set of agents, holds in an ATL  structure if by suitably choosing their moves, the agents in $A$ \emph{can force $\phi$ true}, no matter how other agents happen to move.
The semantics of ATL is defined in so-called \emph{concurrent game structures} where, at each point, all agents simultaneously choose their moves from a finite set, and the next state deterministically depends on such choices.
More concretely, an ATL structure is a tuple 
%\[ 
$
\M = 
\tup{\agents,Q,\P,\actions,\atld,\mapFunc,\transFunc},
$
%\]
where $\agents = \set{1,\ldots,k}$ is a finite set of agents, $Q$ is the finite set of states, $\P$ is the finite set of propositions, $\actions$ is the set of all domain actions, $d : \agents \times Q \mapsto 2^{\actions}$ indicates all available actions for an agent in a state, $\mapFunc : Q \mapsto 2^\P$ is the valuation function stating what is true in each state,  and $\transFunc : Q \times \actions^{\card{\A}} \mapsto Q$ is the transition function mapping a state $q$ and a joint-move $\vec{\act} \in \D(q)$---where $\D(q) = \times_{i=1}^{\card{\A}} d(i,q)$ is the set of legal joint-moves in $q$
---to the resulting next state $q'$.

A \defterm{path} $\lambda = q_0q_1 \cdots$ in a structure $\M$ is a, possibly infinite, sequence of states such that for each $i \geq 0$, there exists a joint-move $\vec{\act_i} \in \D(q_i)$ for which $\transFunc(q_i,\vec{\act_i}) = q_{i+1}$. 
We use $\lambda[i] = q_i$ to denote the $i$-th state of $\lambda$, $\Lambda$ to denote the set of all paths in $\M$, and $\Lambda(q)$ to denote those starting in $q$. 
Also, $|\path|$ denotes the length of $\path$ as the number of state transitions in $\lambda$:  $|\path|=\ell$ if $\path=q_0q_1\ldots q_\ell$, and $|\path|=\infty$ if $\path$ is
infinite. 
When $0 \leq i \leq j \leq |\path|$, then $\path[i,j] = q_i q_{i+1}\ldots q_j$ is the finite subpath between the $i$-th and $j$-th steps in $\lambda$.
Finally, a \defterm{computation path} in $\M$ is an infinite path in $\Lambda$.

To provide semantics to formulas $\coalition{\cdot}\varphi$, ATL relies on the notion of agent strategies. Technically, an ATL \defterm{strategy} for an agent $\agt$ is a function $f_{\agt} : Q^+ \mapsto \actions$, where $f_{\agt}(\lambda  q) \in d(\agt,q)$ for all $\lambda q \in Q^+$, stating a particular action choice of agent $\agt$ at path $\lambda q$. 
A \defterm{collective strategy} for group of agents $A \subseteq \agents$ is a set of strategies $F_A = \set{f_\agt \mid \agt \in \agents}$ providing one specific strategy for each agent $\agt \in A$.
For a collective strategy $F_A$ and an initial state $q$, it is not difficult to define the set $\out(q,F_A)$ of all \defterm{possible outcomes} of $F_A$ starting at state $q$ as the set of all computation  paths that may ensue when the agents in $A$ behave as prescribed by $F_A$, and the remaining agents follow any arbitrary strategy~\cite{AluraHK:JACM02-ATL,Walther:TARK07-ATLES}.
The semantics for the coalition modality is then defined as follows (here $\phi$ is a \emph{path formula}, that is, it is preceded by $\mnext$, $\always$, or $\until$, and $\M,\lambda \models \phi$ is defined in the usual way~\cite{AluraHK:JACM02-ATL}):
\begin{center}
$\M,q \models \coalition{A}\phi$ \emph{iff} there is a collective strategy $F_A$ such that
for all computations $\lambda  \in \out(q, F_A)$, we have $\M,\lambda \models \phi$.
\end{center}

The coalition modality only allows for implicit (existential) quantification over strategies.
In some contexts, though, it is important to refer to strategies explicitly in the language, e.g., can a player win the game if the opponent plays a specified strategy?
To address this limitation, \citea{Walther:TARK07-ATLES} proposed ATLES, an extension of ATL where the coalition modality is extended to $\coalitione{A}{\rho}$, where $\rho$ is a \emph{commitment function}, that is, a partial function mapping agents to so-called \emph{strategy terms}.  
%%
% The idea is that each agent $a \in \agents$ for which $\rho(a)$ is defined, commits to the specific strategy
% $\rho(a)$. 
%%
Formula $\coalition{A}_\rho \phi$ thus means that \emph{``while the agents in the domain of $\rho$ act according to their commitments, the coalition $A$ can cooperate to ensure $\phi$ as an outcome.''}
%%
% To give meaning to the extended coalition modality, the semantics of ATL is extended with a mapping $\parallel\!\cdot\!\parallel$
% from strategy terms to actual ATL strategies.

The motivation for our work stems from the fact that ATLES is agnostic on the source of the strategic terms: all meaningful strategies have already been identified.
In the context of multi-agent systems, it may not be an easy task to identify those strategies compatible with the agents' behaviors, as those systems are generally built using programming frameworks~\cite{Bordinietal:Informatica2006-SurveyLangMAS} that are very different from ATL(ES).

% We observe two things regarding ATLES.
% %%
% First, commitment functions identify agents with \emph{single} (deterministic) strategies.
% %%
% More importantly, the framework is agnostic on the source of the strategic terms; it is assumed that the meaningful strategies have already been identified.
% %%
% In the context of multi-agent systems, it may not be an easy task to identify those strategies compatible with the agents' behaviors, as those systems are generally built using programming frameworks that are very different from ATL(ES).

% In the context of autonomous multi-agent systems, such functions would represent possible behaviors that agents can display. However, it may not be an easy task to identify such functions, since those systems are generally built using programming frameworks that are very different from ATL(ES).

% version includes a specific functionality to \emph{extract} a strategy from a witness.

\subsection{BDI Programming}\label{sec:bdiprog}

The BDI agent-oriented programming paradigm is a popular and successful approach for building agent systems, with roots in philosophical work on rational action~\cite{Bratman:CI88} and a plethora of programming languages and systems available, such as  \JACK, \JASON, \JADEX, \DAPL~\cite{Bordinietal:Informatica2006-SurveyLangMAS}, and \GOALBDI~\cite{Hindriks:Goal07}, among others.

A typical BDI agent continually tries to achieve its goals (or desires) by selecting an adequate plan from its \textit{plan library} given its current beliefs, and placing it into the \textit{intention base} for execution.
The agent's plan library $\Pi$ encodes the standard operational knowledge of the domain by means of a set of \defterm{plan-rules} (or ``recipes'') of the form $\phi[\alpha]\psi$: \textit{plan  $\alpha$ is a reasonable plan to adopt for achieving $\psi$ when (context) condition $\phi$ is believed true}.
For example, walking towards location $x$ from $y$ is a reasonable strategy, if there is a short distance between $x$ and $y$ (and the agent wants to be eventually at location $x$). 
Conditions $\phi$ and $\psi$ are (propositional) formulas talking about the current and goal states, respectively.
Though different BDI languages offer different constructs for crafting plans, most allow for sequences of domain actions that are meant to be directly executed in the world (e.g., lifting an aircraft's flaps), and the posting of (intermediate) \emph{sub-goals} $!\varphi$ (e.g., obtain landing permission) to be resolved. 
The intention base, in turn, contains the current, partially executed, plans that the agent has already \emph{committed to} for achieving certain goals. Current intentions being executed provide a screen of admissibility for attention focus~\cite{Bratman:CI88}.
%~\cite{Bratman:87,Pollack:AIJ92-IRMA}.

% Since sub-goals posted during the execution of a plan are also meant to be resolved via the plan library, the execution of a BDI system can be seen as a \textit{context sensitive subgoal expansion}, allowing agents to ``act as they go'' by making \emph{plan choices} at each level of abstraction with respect to the current situation.
% %%
% Important, also, is the usual plan/goal \emph{failure mechanism} typical of BDI architectures, in which alternative plans for a
% (sub)goal are tried upon failure of the current plan. Plan failure could happen due to an action precondition not holding or the non-availability of plans for a sub-goal. If alternative plans for a goal are not available, then failure is propagated towards higher-level goals/ and plans.

Though we do not present it here for lack of space, most BDI-style programming languages come with a clear single-step semantics basically realizing~\cite{Rao:KR92}'s execution model in which \emph{(rational) behavior arises due to the execution of plans from the agent's plan library so as to achieve certain goals relative to the agent's beliefs}.

\section{\BDIOURS: ATL for BDI Agents}\label{sec:bdiatles}

%A BDI agent executes plans in order to achieve its goals. In ATL and its various extensions there is no notion of agents acting according to their plans. As a result, the existence of a strategy for a coalition is checked according to the physical possibilities available to the agents. However, if the agents are acting according to their plans, then only plan-compatible strategies should be considered. Informally, plan-compatible strategies are such that at every step the agent is executing a plan from its plan library.
%ATLES has strategies explicit in the object language and captures the notion that some agents are committed to executing a set of strategies. The strategies in ATLES are provided as part of the model rather than a product of agent's own deliberation process. In this work, we provide a way to synthesize such strategies such that they are compatible with the agent's goals and plans.

%BDI agent's have capabilities which consist of a set of plans. Capabilities can be seen has having know how to bring about a change in the world. For example, an agent having the capability of fueling a rocket has plans which it can execute to fuel a rocket. Set of all the plans the agent has constitutes its plan library. We start with describing a few technical notions which build up our framework to express plan-compatible strategies.
%

Here we develop an ATL(ES)-like logic that bridges the gap between verification frameworks and BDI agent-oriented programming languages. The overarching idea is for BDI programmers to be able to encode BDI applications in ATL in a principled manner. 
% Also, as ATLES, we want to keep the ability to reason both about what is possible and what is possible relative to certain strategies. 

Recall that ATL(ES) uses strategies to denote the agent's choices among possible actions. For a BDI agent these strategies are \emph{implicit} in her know-how. 
In particular, we envision BDI agents defined with a set of \emph{goals} and so-called \emph{capabilities}~\cite{BusettaRHL:AL99-JACK,Padgham05}. Generally speaking, a capability is a set/module of procedural knowledge (i.e., plans) for some functional requirement. An agent may have, for instance,  the \Navigation capability encoding all plans for navigating an environment. Equipped with a set of capabilities, a BDI agent  executes actions as per plans available so as to achieve her goals, e.g., exploring the environment. 
In this context, the BDI developer is then interested in what agents can achieve at the level of goals and capabilities. 
Inspired by ATLES, we develop a logic that caters for this requirement without departing much from the ATL framework.

% As described later, a capability maps to a family of strategies rather than just one, and generating all of these by hand is a tedious task and not humanely possible. The ATLES framework uses the \emph{commitment function} to express that some agents are committed to following a particular strategy. In this setting, we modify the ATLES framework to allow reasoning based on capabilities rather than strategies. We chose to base our framework on ATLES since it allows defining commitment of agents to strategies that are based on histories rather than being memory-less.

\newcommand{\allplans}{\pmb\Pi}
In this work, we shall consider plans consisting of single actions, that is, given BDI plan for the form $\plan{}$, the body of the plan $\alpha$ consists of one primitive action. Such plans are akin to those in the \GOALBDI agent programming language~\cite{Hindriks:Goal07}, as well as universal-plans~\cite{Schoppers:IJCAI87-UniversalPlans}, and reactive control modules~\cite{Baral:ETAI98}. 
Let $\allplans^{\P}_{\actions}$ be the (infinite) set of all possible plan-rules given a set of actions $\actions$ and a set of domain propositions $\P$.

% Given an agent $\agt$ with capabilities $c_1,\ldots ,c_n$, the set of strategies induced by its capabilities is denoted by $\Sigma^{\agt}_{\Pi}$, where $\Pi = \union_{i\leq n}\Pi^{c_i}$. Here, $\Pi$ represents the plan library of the agent, and we call such strategies to be plan-compatible. Intuitively, plan-compatible strategies are such that they result in traces such that the agent is always executing plans to achieve its goals. We formally define which strategies constitute the set $\Sigma^{\agt}_{\Pi}$ after describing the required modifications to ATLES to represent the capabilities.

%In order to modify ATLES to represent BDI-agents, we replace the commitment function $\rho$ of ATLES with capability function $\tau$. The capability function is a partial function mapping an agent to the set of capabilities they use. That is, each $\agt \in dom(\tau)$ acts according to its capabilities $\tau(\agt)$. Similarly, we replace the coalition modality by $\coalition{}_\tau$ where $\tau$ is the capability function. Then $\coalition{A}_\tau \varphi$ means that while the agents in $dom(\tau)$ are acting according to their capabilities the coalition A can enforce $\varphi$. We call the resulting modification as \emph{BDI-ATLES}.

% \vspace*{-.3cm}
\subsection{BDI-ATLES Syntax}

The language of \BDIOURS is defined over a finite set of atomic propositions $\P$, a finite set of agents $\agents$, and a finite set of capability terms $\C$ available in the BDI application of concern. Intuitively, each capability term $c \in \C$ (e.g., $\Navigation$) stands for a plan library $\Pi^{c}$ (e.g., $\Pi^{\Navigation}$). 
As usual, a \defterm{coalition} is a set $A \subseteq \agents$ of agents. 
A \defterm{capability assignment} $\omega$ consists of a set of pairs of agents with their capabilities of the form $\tup{\agt:C_{\agt}}$, where $\agt \in \agents$ and $C_\agt \subseteq \C$. 
A \defterm{goal assignment} $\varrho$, in turn, defines the goal base (i.e., set of propositional formulas) for some agents, and is a set of tuples of the form  $\tup{\agt:G_\agt}$, where $\agt \in \agents$ and $G_\agt$ is a set of boolean formulas over $\P$.
We use $\agents_{\omega}$ to denote the set of agents for which their capabilities are defined by assignment $\omega$, that is, $\A_{\omega} = \set{\agt \mid \tup{\agt:C_\agt} \in \omega}$. Set $\A_{\varrho}$ is defined analogously.
%%
% Given a capability assignment $\omega$ (resp., goal assignment $\varrho$), we denote $\agents_{\omega} \subseteq \agents$ (resp., $\A_{\varrho} \subseteq \agents$) the set of agents for which their capabilities (resp., goal bases) are defined by assignment $\omega$ (resp., $\varrho$), that is, $\A_{\omega} = \set{\agt \mid \tup{\agt:C_\agt} \in \omega}$ (resp., $\A_{\varrho} = \set{\agt \mid \tup{\agt:G_\agt} \in \varrho}$).

% The set of \BDIOURS formulas is defined by the following grammar:
% \[
% \varphi:=p|\neg \varphi|\varphi_1 \vee \varphi_2|\coalition{A}_{\omega,\varrho}\mnext\varphi|\coalition{A}_{\omega,\varrho}\always\varphi|\coalition{A}_{\omega,\varrho}\varphi_1\until\varphi_2,
% \]
% %%%
% where $p \in \P$, $A$ ranges over coalitions, $\omega$ and $\varrho$ range over capability and goal assignments such that $\agents_{\omega}=\agents_{\varrho}$, respectively, and $\varphi$ ranges over BDI-ATLES formulas. 
% %%
% Note that the syntax is very much that of ATLES, except that capability and goal assignments are used instead of commitment functions.

%%
The set of \BDIOURS formulas is then exactly like that of ATL(ES), except that coalition formulas are now of the form $\coalition{A}_{\omega,\varrho}\varphi$, where $\varphi$ is a path formula (i.e., it is preceded by $\mnext$, $\always$, or $\until$), $A$ is a coalition, and $\omega$ and $\varrho$ range over capability and goal assignments, respectively, such that $\agents_{\omega}=\agents_{\varrho}$. 
%%
% Note that the syntax is very much that of ATLES, except that capability and goal assignments are used instead of commitment functions.
%%
Its intended meaning is as follows:

\begin{quote}
$\coalition{A}_{\omega,\varrho}\varphi$ expresses that coalition of agents $A$ can jointly force temporal condition $\varphi$ to hold when BDI agents in $\A_\omega$ (or $\A_\varrho$, since $\A_\varrho=\A_\omega$) are equipped with capabilities as per assignment $\omega$ and (initial) goals are per assignment $\varrho$.
\end{quote}

Notice that we require, in each coalition (sub)formula, that the agents for which capabilities and goals are assigned to be the same. This enforces the constraint that BDI-style agents have \emph{both} plans and goals.
Hence, a formula of the form $\coalition{A}_{\emptyset,\set{\tup{a_1:\set{\gamma}}}} \varphi$ would not be valid, as agent $a_1$ has one goal (namely, to bring about $\gamma$), but its set of plans is not defined---we cannot specify what its rational behavior may be. This contrasts with formula $\coalition{A}_{\set{\tup{a_1:\emptyset}},\set{\tup{a_1:\set{\gamma}}}}\varphi$, a valid formula in which agent $a_1$ is assumed to have no plans (i.e., agent has empty know-how) and one goal.
% 
% The reason is simple: a BDI agent is specified with BOTH goals AND plans. If only one component is specified, it is unclear what the right behavioral interpretation should be. (See not specified is different from empty.)

%%%%%%%%%%%%%%%%%% EXAMPLE

\begin{example}
Consider the following simplified instance of the gold mining domain with three locations $A$, $B$ and $C$, a gold piece $\diamond$ at location $C$, the depot located at $B$ (rectangle location), and two players $\Ag$ (BDI agent) and $\En$ (enemy):
{\small
\begin{center}
% {\resizebox{\textwidth}{!}{
\vspace*{-.3cm}
\begin{tikzpicture}[-,>=stealth',shorten >=1pt,auto,node distance=2cm,semithick,scale=.5]
\tikzstyle{every state}=[circle,fill=none,draw=black,text=black,inner sep=0pt,minimum size=4mm,font=\scriptsize, text centered]

\node[state]    (q0) [label=left:$A$]         		{$\En$};
\node[state,rectangle]    (q1) 
	[right of=q0,label=5:$B$] 	    {$\Ag$};
\node[state]    (q2) [right of=q1,label=right:$C$]      
	{$\diamond$};

\path
	(q0) edge               node {} (q1)
    (q1) edge               node {} (q2)
;
\end{tikzpicture}

% }
% }
\vspace*{-.1cm}
\end{center}
}
Players can move $\aLeft$/$\aRight$, $\aPick$/$\aDrop$ gold, or remain still by executing special action $\nop$.
Proposition $X_Y$, where $X \in \set{\Ag,\En}$ and $Y\in \set{A,B,C}$, encodes that player $X$ is at location $Y$; whereas propositions $\gA$, $\gB$, $\gC$, $\gAg$, and $\gEn$ denote that the gold is at location $A$/$B$/$C$ or being held by agent $\Ag$/$\En$, respectively. 
%%
% \begin{itemize}\separation{-.05in}
%   \item $X_Y$, with $X \in \set{\Ag,\En}$ and $Y\in \set{A,B,C}$, states that player $X$ is at location $Y$.
%    
%   \item $\gA$, $\gD$, $\gC$, $\gAg$, and $\gEn$ denote that locations $A$, $B$, $C$, agents $\Ag$ and $\En$, respectively, have gold.
%   
%   
%   \item $W_X$, with $X \in \set{\Ag,\En}$ states that player $X$ has won.
% \end{itemize}
%
The depot is assumed to be always at $B$ and hence is not represented explicitly.
%%
% If the gold is dropped at location $B$ it goes into the depot, therefore we do not use any proposition to denote the location of gold at $B$.

The winning condition for player $\Ag$ is $\psi_{\WIN} \!\! =\! \gB \land \agB$: the player wins when collocated with gold at the depot.

Among the many capabilities available encoding the know-how information of the domain, we consider the following three.
%%
% One could imagine several capabilities encoding the know-how information of the domain.
%%
The $\Pick$ capability includes plans to pick gold, such as $\agC\wedge\gC[\aPick]\gB$: if gold needs to be at $B$ and agent is at $C$, where there is indeed gold, then execute the $\aPick$ action. 
Similarly, capability $\Drop$ contains plans like $\gAg\wedge\agB[\aDrop]\gB$, for example, to allow dropping of gold at the desired location.
Lastly, capability $\Navigation$ has plans for moving around, such as $\agC[\aLeft]\agB$ to move left from location $C$ to (desired destination) $B$.
\qed
\end{example}

The remaining of the section involves providing the right interpretation to such formulas, under the assumption that agents act rationally as per the BDI paradigm.

\subsection{BDI-ATLES Semantics}

A \BDIOURS \defterm{concurrent game structure} is a tuple $\M  \!\!\! = \!\!\!  \tup{\agents,Q,\P,\actions,\atld,\mapFunc,\transFunc,\mapCap}$, with:
\begin{itemize} %\setlength{\itemsep}{-.75mm}
  \item $\agents$, $Q$, $\P$, $\actions$, $\atld$, $\mapFunc$ and $\transFunc$ are as in ATL(ES).
      
  \item There is a distinguished dummy action $\nop \in \actions$
  such that $\nop \in d_{\agt}(q)$ and $\sigma(q,\tup{\nop,\ldots,\nop}) = q$, for all $\agt \in \agents$ and $q \in Q$, that is, $\nop$ is always available to all agents and the system remains still when all agents perform it.
  
\item Capability function $\mapCap: \C \mapsto {\F(\allplans^{\P}_{\actions})}$ maps capability terms to their (finite) set of plans. (Here, $\F(X)$ denotes the set of all finite subsets $X$.)
\end{itemize}
 
%%%%%%%%%%%%%%%%%%%%%%%%%%% EXAMPLE %%%%%%%%%%%%%%%%%%%
\begin{figure*}[t]
\centering
\subfigure[A section of the \BDIOURS alternating model.]{\label{fig:goldModel}
\resizebox{0.52\textwidth}{!}{\begin{tikzpicture}[->,>=stealth',shorten >=1pt,auto,node distance=2.5cm,semithick]
\tikzstyle{every state}=[circle ,fill=none,draw=black,text=black,font=\scriptsize, text centered,inner sep=0pt, minimum size=2pt]
\begin{scope}[]
%good trace
\node[state]    (q0) [label=above:$q_0$]         
	{\hspace{-0.0cm} 
		\begin{tabular*}{1.3cm}{c}
        	 $\agB$, $\enA$ \\ 
        	% $\lnot \WAg$, $\lnot \WEn$ \\  
        	 $\gC$
		\end{tabular*}
	};

\node[state]    (q1) [right of=q0, label=above:$q_1$] 	    {\hspace{0.0cm}\begin{tabular*}{1.3cm}{c}
                                                            $\agC$, $\enB$ \\ 
                                                            %$\lnot \WAg$, $\lnot \WEn$ \\  
                                                            $\gC$
                                                            \end{tabular*}};

\node[state]    (q2) [right of=q1, label=above:$q_2$]        {\hspace{-0.0cm}\begin{tabular*}{1.3cm}{c}
                                                           $\agC$, $\enC$ \\ 
                                                           %$\lnot \WAg$, $\lnot \WEn$ \\
                                                             $\gAg$
                                                            \end{tabular*}};
                                                            
\node[state]    (q3) [right of=q2, label=above:$q_3$]        {\hspace{-0.0cm}\begin{tabular*}{1.3cm}{c}
                                                            $\agB$, $\enC$ \\ 
                                                            %$\lnot \WAg$, $\lnot \WEn$ \\
                                                              $\gAg$
                                                            \end{tabular*}};
                                                            
\node[state]    (q4) [right of=q3, label=above:$q_4$]        {\hspace{-0.0cm}\begin{tabular*}{1.3cm}{c}
                                                            $\agB$, $\enC$ \\ 
                                                            %$\WAg$, $\lnot \WEn$\\
                                                              $\gB$
                                                            \end{tabular*}};

% lacking pick capability
\node[state]    (q5) [below of=q1, yshift=.2cm,label=below:$q_5$]        {\hspace{-0.0cm}\begin{tabular*}{1.3cm}{c}
                                                            $\agC$, $\enC$ \\
                                                             %$\lnot \WAg$, $\lnot \WEn$ \\
                                                               $\gC$
                                                            \end{tabular*}};
                                                            
\node[state]    (q6) [right of=q5,label=below:$q_6$]        {\hspace{-0.0cm}\begin{tabular*}{1.3cm}{c}
                                                            $\agC$, $\enC$ \\
                                                             %$\lnot \WAg$, $\lnot \WEn$ \\
                                                               $\gEn$
                                                            \end{tabular*}};  
                                                            
\node[state]    (q7) [right of=q6, label=below:$q_7$]        {\hspace{-0.0cm}\begin{tabular*}{1.3cm}{c}
                                                            $\agC$, $\enB$ \\
                                                             %$\lnot \WAg$, $\lnot \WEn$ \\
                                                               $\gEn$
                                                            \end{tabular*}};
                                                            
\node[state]    (q8) [right of=q7, label=below:$q_8$]        {\hspace{-0.0cm}\begin{tabular*}{1.3cm}{c}
                                                            $\agC$, $\enB$ \\
                                                             %$\lnot \WAg$, $\WEn$ \\
                                                               $\gB$
                                                            \end{tabular*}};                                           
  
\node[state]    (q9) [left of=q5, label=below:$q_9$]        {\hspace{-0.0cm}\begin{tabular*}{1.3cm}{c}
                                                            $\agB$, $\enB$ \\ 
                                                            %$\lnot \WAg$, $\lnot \WEn$ \\
                                                              $\gC$
                                                            \end{tabular*}};                                                                                                      
                                                                               
% info node
%\node[] (i2) [below of=q5,yshift=0.75cm,xshift=1.5cm] {Actions $:=\{ l:left, r:right, p:pick, d:drop, n:\nop \}$};

\path
	(q0) edge               node {$\tup{r,r}$} (q1)
    (q1) edge               node {$\tup{p,r}$} (q2)
    (q2) edge               node {$\tup{l,n}$} (q3)
    (q3) edge               node {$\tup{d,n}$} (q4)
    (q0) edge[loop left, distance=0.75cm]   
    		node[below] 
    		{\begin{tabular}{c}
    			$\tup{n,n}$ \\
    			$\tup{p,p}$ \\
    			%$\tup{d,d}$
    		\end{tabular}} (q0)
% lacking pick capability
    (q1) edge[bend left]    node {$\tup{n,r}$} (q5)
    (q5) edge               node {$\tup{n,p}$} (q6)
    (q6) edge               node {$\tup{n,l}$} (q7)
    (q7) edge               node {$\tup{n,d}$} (q8)

% lacking drop capability
    (q2) edge[loop below, distance=0.75cm]   
    	node[right] {$\tup{n,n}$} (q2)

% not rational
    (q0) edge[bend left]    node[near start] {$\tup{n,r}$} (q9)
    (q9) edge               node {$\tup{r,r}$} (q5)
    (q5) edge[bend left]   node	{$\tup{n,l}$} (q1)
    (q9) edge[bend left ]   node[near start] {$\tup{n,l}$} (q0)    
;
\end{scope}
\end{tikzpicture}}
}
\quad
\subfigure[Traces $\path^+_1$ and $\path^+_2$ resultant from strategies $f^1_{\Ag}$ and $f^2_{\Ag}$, respectively]{\label{fig:traces}
\resizebox{0.42\textwidth}{!}{\begin{tikzpicture}[->,>=stealth',shorten >=1pt,auto,node distance=2cm,semithick]
\tikzstyle{every state}=[circle,fill=none,draw=black,text=black,inner sep=1pt,minimum size=6mm,font=\scriptsize]

% root node
\node[state]    (A) []			{$q_0$};

% trace 1 states
\node[state]    (B1) [above right of=A,yshift=-0.0cm] {$q_1$};
\node[state]    (C1) [right of=B1]		{$q_2$};
\node[state]    (D1) [right of=C1]		{$q_3$};
\node[state]    (E1) [right of=D1] 		{$q_4$};

% trace 2 states
\node[state]    (B2) [below right of=A,yshift=0.5cm] {$q_9$};
\node[state]    (C2) [right of=B2] 		{$q_5$};
\node[state]    (D2) [right of=C2] 		{$q_1$};
\node[state]    (E2) [right of=D2] 		{$q_2$};

% info node
%\node[] (i) [below of=A,yshift=-0.75cm,xshift=1.5cm] {$\Pi=\{r[abc]p,\ s[xy]p\}$, $c \not\in d(\agt,q_2)$.};

% trace labels
\node[] (l1) [right of=E1,xshift=-1.45cm] {$\lambda^+_1$};
\node[] (l2) [right of=E2,xshift=-1.45cm] {$\lambda^+_2$};

\node[] (f1) [above of=E1,xshift=0cm,yshift=-1.25cm] {$f^1_{\Ag} \in \Sigma^{\Ag}_{\Pi,\G}$};
\node[] (f2) [above of=E2,xshift=0cm,yshift=-1.25cm] {$f^2_{\Ag} \not\in \Sigma^{\Ag}_{\Pi,\G}$};

% markings
\draw[-, rounded corners=2pt] (-0.28,0.38)--(-0.4,0.5)--(1.0,1.9)--(1.18,1.80);
\node[] (p1) [above of=A,xshift=-0.05cm,yshift=-0.75cm,rotate=45] {$\agB \wedge \gC[r]\gD$};

\draw[-, rounded corners=2pt] (1.4,1)--(1.4,0.8)--(3.4,0.8)--(3.4,1);
\node[] (p1) [below of=C1,xshift=-1.1cm,yshift=1cm] {$\agC \wedge \gC[p]\gD$};

\draw[-, rounded corners=2pt] (3.4,1.78)--(3.4,2.0)--(5.4,2.0)--(5.4,1.75);
\node[] (p1) [above of=C1,xshift=1cm,yshift=-1.15cm] {$\agC \wedge \gAg[l]\agB$};

\draw[-, rounded corners=2pt] (5.4,1)--(5.4,0.8)--(7.3,0.8)--(7.3,1);
\node[] (p1) [below of=E1,xshift=-1.1cm,yshift=1cm] {$\agB \wedge \gAg[d]\gD$};

\path
% trace 1 actions
	(A) edge               node {$r$} (B1)
	(B1) edge              node {$p$} (C1)
 	(C1) edge              node {$l$} (D1)
 	(D1) edge              node {$d$} (E1)
%trace 2 actions
 	(A) edge               node {$n$} (B2)
	(B2) edge              node {$r$} (C2)
 	(C2) edge              node {$n$} (D2)
 	(D2) edge              node {$p$} (E2)
;
\end{tikzpicture}}
}
\caption{
% \textit{(a)} Simplified Gold domain instance with two agents $\Ag$ and $\En$; gold ($\diamond$), which is present at location $C$, is to be deposited in the depot ($\bullet$) at location $B$; 
%%
A fragment of a Gold domain model and a picture showing rational traces and strategies.
Actions $\aLeft$, $\aRight$, $\aPick$, $\aDrop$, and $\nop$ are abbreviated with their first letter.
}
\end{figure*}
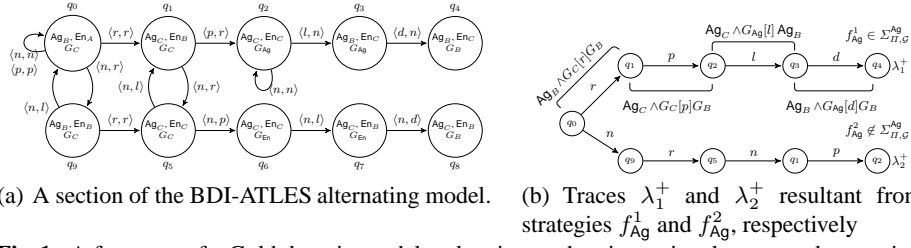

% 
% \begin{figure*}[t]
% \begin{center}
% \input{fig-goldModel}
% \end{center}
% \caption{A simplified Gold domain model (partial). Actions $:=\set{l:\aLeft, r:\aRight, p:\aPick, d:\aDrop, n:\nop}$}
% \label{fig:goldModel}
% \end{figure*}

\begin{example}
Figure~\ref{fig:goldModel} shows a partial model for the gold game.
The game starts at state $q_0$, with players $\Ag$ and $\En$ located at $B$ and $A$, resp., and gold present at $C$. 
From there, player $\Ag$ has a winning strategy: reach the gold earlier and deposit it in the depot.
This can be seen in path $q_0q_1q_2q_3q_4$. However, this is possible only when the agent $\Ag$ is indeed equipped with all  three capabilities.
If, on the other hand, the agent lacks capability $\Pick$, for instance, then player $\En$ may actually manage to win the game, as evident from the path $q_0q_1q_5q_6q_7q_8$. 
\qed
\end{example}

\BDIOURS models are similar to ATLES ones, except that capability, rather than strategy term, interpretations are used.
In a nutshell, the challenge thus is to characterize what the underlying ``low-level'' ATL strategies for agents with certain capabilities and goals are. We call such strategies \emph{rational strategies}, in that they are compatible with the standard BDI rational execution model~\cite{Rao:KR92}: \emph{they represent the agent acting as per her available plans in order to achieve her goals in the context of her beliefs}.

So, given an agent $\agt \in \agents$, a plan-library $\Pi$, and a goal base $\G$, we define $\Sigma^{\agt}_{\Pi,\G}$ to be the set of standard ATL strategies for agent $\agt$ in $\M$ that are \defterm{rational strategies} when the agent is equipped with plan-library $\Pi$ and has $\G$ as (initial) goals, that is, those ATL strategies in which the agent always chooses an action that is directed by one of its available plans in order to achieve one of its goals in the context of its current beliefs.
The core idea behind defining set $\Sigma^{\agt}_{\Pi,\G}$ is to identify those ``rational traces'' in the structure that are compatible with the BDI deliberation process in which the agent acts as per her goals and beliefs. 
Traces just generalize paths to account for the actions performed at each step, and are hence of the form $\lambda^+=q_0\act_1 q_1\cdots\act_\ell q_\ell$ such that $q_0q_q\cdots q_\ell$ is a (finite) path.
Rational strategies, then, are those that only yield rational traces. 
Technically, we define \emph{rational traces} in three steps. 
First, we define a \emph{goal-marking} function $\goalFunc(\lambda^+,i)$ denoting the ``active'' goal base of the agent at the $i$-th stage of trace $\lambda^+$. Basically, a goal-marking function keeps track of the goals that the agent has already achieved at each stage in a trace. 
Second, we define  $\exec(\plan{},\goalFunc,\lambda^+)$ as the set of indexes (i.e., stages) in trace $\lambda^+$ where the plan $\plan{}$ may have been executed by the agent: the plan's precondition $\phi$ was true, $\psi$ was an active goal of the agent (as directed by goal-marking function $\goalFunc$), and $\alpha$ was indeed performed. 
Finally, we say a trace $\lambda^+$ is deemed ``rational'' if at every moment in the run the agent executed one of its plans. That is, for every index $i$, it is the case that $i \in \exec_\agt(\plan{},\goalFunc,\lambda^+)$, for some plan $\plan{}$ in her know-how library. 
Finally, we use $\Sigma^{\agt}_{\Pi,\G}$ to denote the set of all ATL strategies whose executions always yield rational traces. The laborious, and arguably boring, technical details of all this can be found in the Appendix.

\begin{example}
Figure~\ref{fig:traces} depicts two possible traces $\path^+_1$ and $\path^+_2$ (for agent $\Ag$) compatible with strategies $f^1_{\Ag}$ and $f^2_{\Ag}$, resp.
%%
%The agent $\Ag$ has the initial goal of $\WAg$ and is equipped with all the three capabilities.
%the $\Navigation$, $\Pick$ and $\Drop$ capabilities.
%%
Trace $\path^+_1$ is due to the agent executing actions as per its applicable plans, as evident from the plan labeling. For example, at state $q_1$, the agent is in a gold location and hence executes the pick action as per plan $\agC\wedge\gC[\aPick]\gB$. Consequently the strategy $f^1_{\Ag}$ is rational, as it yields rational trace $\path^+_1$.
Trace $\path^+_2$ on the other hand does not obey the BDI rationality constraints (e.g., the agent remains still in location $B$, despite an applicable plan being available).
\qed 
\end{example}

Assuming that set $\Sigma^{\agt}_{\Pi,\G}$ of rational strategies has been suitably defined, we are ready to detail the semantics for formulas of the form $\coalitione{A}{\omega,\varrho}\varphi$.
Following ATLES we first extend the notion of a joint strategy for a coalition to that of joint strategy \emph{under a given capability and goal assignment}.
So, given a  capability (goal) assignment $\omega$ ($\varrho$) and an agent $\agt \in \A_{\omega}$ ($\agt \in \A_{\varrho}$), we denote $\agt$'s capabilities (goals) under $\omega$ ($\varrho$) by $\omega[\agt]$ ($\varrho[\agt]$).
Intuitively, an \emph{\STRATEGYOURS-strategy for coalition $A$} is a joint strategy for $A$ such that \emph{(i)} agents in $A\cap \agents_{\omega}$ only follow ``rational'' (plan-goal compatible) strategies as per their $\omega$-capabilities and $\varrho$-goals; and \emph{(b)} agents in $A\backslash \agents_{\omega}$ follow arbitrary strategies. 
Formally, an \defterm{\STRATEGYOURS-strategy for coalition $A$} (with $\A_\omega=\A_\varrho$) is a collective strategy $F_A$ for agents $A$ such that for all $f_{\agt} \in F_A$ with $\agt \in A\cap \agents_{\omega}$, it is the case that $f_{\agt} \in \Sigma^{\agt}_{\Pi,\G}$, where $\Pi = \cup_{c \in \omega[\agt]}\mapCap(c)$ and $\G = \varrho[\agt]$. Note no requirements are asked on the strategies for the remaining agents $A\backslash \agents_{\omega}$, besides of course being legal (ATL) strategies. 
Also, whereas ATLES $\rho$-strategies are collective strategies including \emph{all} agents in the domain of commitment function $\rho$, our \STRATEGYOURS-strategies are collective strategies for the coalition of concern only. This is because commitment functions induce deterministic agent behaviors, whereas capabilities and goals assignments induce non-deterministic ones. 
We will elaborate on this issue below.

Using the notions of \STRATEGYOURS-strategies and that of possible outcomes for a given collective strategy from ATL (refer to function $\out(\cdot,\cdot)$ from Preliminaries), we are now able to state the meaning of \BDIOURS (coalition) formulas:%
\footnote{As with ATL(ES), $\varphi$ ought to be a path formula and is interpreted in the usual manner. We omit the other ATL-like cases for brevity; see~\cite{Walther:TARK07-ATLES}.}
%%
% Using the notion of \STRATEGYOURS-strategies and that of possible outcomes for a given collective strategy from ATL (refer to function $\out(\cdot,\cdot)$ from Preliminaries), we are now able to state the meaning of \BDIOURS formulas, in particular of coalition formulas:\footnote{Again, as with ATL(ES), formula $\varphi$ ought to be a path formula and it is interpreted in the usual manner. We omit the other ATL-like cases for brevity; see~\cite{Walther:TARK07-ATLES,AluraHK:JACM02-ATL}}
%%
\begin{center}
$\M,q \models \coalition{A}_{\omega,\varrho} \varphi$ \emph{iff} there is a $\tup{\omega,\varrho}$-strategy $F_A$ such that for all $\tup{\omega,\varrho}$-strategies $F_{\A_{\omega}\setminus A}$ for $\A_\omega \setminus A$,  it is the case that $\M,\path\models\varphi$, for all paths $\path \in \outf(q,F_A \cup F_{\A_{\omega}\setminus A})$.
\end{center}

\noindent
Intuitively, $F_A$ stands for the collective strategy of agents $A$ guaranteeing the satisfaction of formula $\varphi$. Because $F_A$ is a \STRATEGYOURS-strategy, some agents in $A$---those whose capabilities and goals are defined by $\omega$ and $\varrho$, resp.---are to follow strategies that correspond to  rational executions of its capabilities.
At the same time, because other agents outside the coalition could have also been assigned capabilities and goals, the chosen collective strategy $F_A$ needs to work no matter how such agents (namely, agents $\A_\omega \setminus A$) behave, as long as they do it rationally given their plans and goals. That is, $F_A$ has to work with \emph{any} rational collective strategy $F_{\A_{\omega}\setminus A}$. 
Finally, the behavior of all remaining agents---namely those in $\A \setminus (A \cup \A_\omega)$---are taken into account when considering all possible outcomes, after all strategies for agents in $A \cup \A_\omega$ have been settled.

While similar to ATLES coalition formulas $\coalitione{A}{\rho}\varphi$, \BDIOURS coalition formulas $\coalitione{A}{\omega,\varrho} \varphi$ differ in one important aspect that makes its semantics more involved. 
Specifically, whereas commitment functions $\rho$ prescribe \emph{deterministic} behaviors for agents, capabilities and goals assignments yield multiple potential behaviors for the agents of interest. This nondeterministic behavior stems from the fact that BDI agents can choose what goals to work on at each point and what available plans to use for achieving such goals.
Technically, this is reflected in the strategies for each agent in $(\A_\omega\setminus A)$---those agents with assigned capabilities and goals but  not part of the coalition---cannot be (existentially) considered together with those of agents in $A$ or (universally) accounted for via the possible outcomes function $\out(\cdot,\cdot)$, as such function puts no rationality constraints on the remaining (non-committed) agents.
Thus, whereas agents in $A \cap \A_\omega$ are allowed to select one possible rational behavior, all rational behaviors for agents in $(A_\omega \setminus A)$ need to be taken into consideration.

We close this section by noting an important, and expected, monotonicity property of $\BDIOURS$ w.r.t. changes in the goals and plans of agents. 

\begin{proposition}
$\models \coalitione{A}{\omega,\varrho} \varphi  \supset \coalitione{A'}{\omega',\varrho'}\varphi$ holds, provided that:
 \begin{itemize}
   %% COALITION MONOTONICITY
   \item $A \subseteq A'$, that is, the coalition is not reduced;

   %% GOAL MONOTONICITY
   \item  $\omega[\agt] \subseteq \omega'[\agt]$ and $\varrho[\agt] \subseteq \varrho'[\agt]$, for all $\agt \in \A_{\omega} \cap A$, that is, the goals and capabilities of those BDI agents in the coalition are not reduced; and

   \item $\A_\omega \setminus A \subseteq \A_{\omega'} \setminus A'$, that is, the set of non BDI agents outside the coalition is not reduced (but could be new BDI agents outside the coalition);
   
   \item $\omega'[\agt] \subseteq \omega[\agt]$ and $\varrho'[\agt] \subseteq \varrho[\agt]$, for all $\agt \in \A_{\omega} \setminus A$, that is, the goals and capabilities of those BDI agents outside the coalition are not augmented.
 \end{itemize}

% \begin{itemize}
%   \item $\models \coalitione{A}{\omega,\varrho} \varphi  \supset \coalitione{A}{\omega',\varrho}\varphi$ if 
%   $\A_{\omega} = \A_{\omega'}$, $\omega[\agt] \subseteq \omega'[\agt]$ for all $\agt \in \A_{\omega} \cap A$ and $\omega[\agt] \not\subset \omega'[\agt]$ for all $\agt \in \A_{\omega} \setminus A$.
%   %%
%   \item $\models \coalitione{A}{\omega,\varrho} \varphi  \supset \coalitione{A}{\omega,\varrho'}\varphi$ if 
%   $\A_{\varrho} = \A_{\varrho'}$, $\varrho[\agt] \subseteq \varrho'[\agt]$ for all $\agt \in \A_{\varrho} \cap A$ and $\varrho[\agt] \not\subset \varrho'[\agt]$ for all $\agt \in \A_{\omega} \setminus A$.
% \end{itemize}
\end{proposition}

Informally, augmenting the goals/plans of agents in a coalition does not reduce the ability of agents.
This is because a collective $\tup{\omega,\varrho}$-strategy for coalition $A$ to bring about a formula would still work if more goals and plans are given to the agents in the coalition (second condition). 
Observe, on the other hand, that augmenting the goals or plans of those agents outside the coalition may yield new behavior that can indeed interfere with the coalition's original abilities (last condition). This even includes turning BDI agents into non BDI agents (third condition).
Of course, as in ATL, enlarging the coalition does not reduce ability (first condition).

\section{BDI-ATLES Model Checking}\label{sec:modelChecking}

\newcommand{\iamahere}{
\begin{center}
\textbf{************* I AM HERE ************** }
\end{center}
}

\newcommand{\ws}{\text{\emph{ws}}}
\newcommand{\goal}{\text{\emph{gl}}}
\newcommand{\BDIATLESCOLFORM}{\coalitione{A}{\omega,\varrho}\varphi\xspace}
\newcommand{\ATLCOLFORM}{\coalition{A}\varphi\xspace}
\newcommand{\Pre}{\text{\emph{Pre}}}
\newcommand{\Rational}{d^{\text{BDI}}}
\newcommand{\applicable}{\text{\emph{appl}}}
\renewcommand{\applicable}{\Delta}

\newcommand{\sbrack}[1]{[#1]}
\newcommand{\dbrack}[1]{\llbracket #1 \rrbracket}

Given a BDI-ATLES model $\M$ and a formula $\varphi$, the model checking algorithm for BDI-ATLES computes the set of states in $\M$ that satisfy $\varphi$.
To that end, the algorithm has to take into account the rational choices of each BDI agent, that is, those choices that are the consequence of the agent's goals and capabilities specified by functions $\varrho$ and $\omega$ in formulae of the form $\BDIATLESCOLFORM$.
Roughly speaking, the algorithm restricts, at each step, the options of BDI agents to their applicable plans.
We start by extending the model $\M$ to embed the possible goals (based on the goal assignment) of BDI agents into each state, and then then discuss the model checking algorithm and its complexity.

%%%%%%%%%%%%%%%% BDI-ATLES Symbolic Model Checking algorithm %%%%%%%%%%%%%%%%
{\small
%\SetKw{od}{od;}
%\begin{algorithm}
%\ForEach{ $\varphi'$ in Sub($\varphi$)}
%{
%    \lCase{$\varphi'=p:$ }{$[\varphi']=\mapFunc(p)$}
%    \\
%    \lCase{$\varphi'= \lnot \theta:$}{$[\varphi']=[true]\setminus[\theta]$}
%    \\
%    \lCase{$\varphi'=\theta_1 \vee \theta_2:$}{$[\varphi'] = [\theta_1] \union [\theta_2]$}
%    \\
%    \lCase{$\varphi' = \coalitione{A}{\omega,\varrho}\mnext \theta:$}{
%        $[\varphi'] = Pre(A,\omega,[\theta]) \cap \llbracket \varrho \rrbracket$
%    }
%    \\
%    \Case{$\varphi' = \coalitione{A}{\omega,\varrho}\always \theta:$}{
%        $\rho=[true]; \tau=[\theta]$ \;
%        \While{$\rho \not\subseteq \tau$}{$\rho=\tau; \tau = Pre(A,\omega,\rho) \cap [\theta]$}
%        $[\varphi']=\rho \cap \llbracket \varrho \rrbracket$
%    }
%    \Case{$\varphi' = \coalitione{A}{\omega,\varrho} \theta_1 \until \theta_2:$}{
%        $\rho=[false]; \tau=[\theta_2]$ \;
%        \While{$\tau \not\subseteq \rho$}{$\rho=\rho \union \tau; \tau = Pre(A,\omega,\rho) \cap [\theta_1]$}
%        $[\varphi']=\rho \cap \llbracket \varrho \rrbracket$
%    }
%}
%\od\\
%\Return $[\varphi']$

%\caption{BDI-ATLES symbolic model checking}
%\end{algorithm}

% \SetAlFnt{\small}
\setlength{\algomargin}{0em}
\SetKw{od}{od;}
\SetKw{ecase}{end case}
\begin{algorithm}[t]
\ForEach{ $\varphi'$ in Sub($\varphi$) w.r.t. $\M = \tup{\agents,Q,\P,\actions,\atld,\mapFunc,\transFunc,\mapCap}$}
{
    %%% CASE 1 %%%%%
    \lCase{$\varphi'=p:$ }
        {$\sbrack{\varphi'}_{\M}=\mapFunc(p)$};
    \\
    %%% CASE 2 %%%%%
    \lCase{$\varphi'= \lnot \theta:$}
        {$\sbrack{\varphi'}_{\M} = (\sbrack{\true}_{\M} \setminus \sbrack{\theta}_{\M})$};
    \\
    %%% CASE 3 %%%%%
    \lCase{$\varphi'=\theta_1 \vee \theta_2:$}
        {$ \sbrack{\varphi'}_{\M} = \sbrack{\theta_1}_{\M} \union \sbrack{\theta_2}_{\M}$};
    \\
    %% coalition formulae start here %%
    %%% CASE 4 %%%%%
    \lCase{$\varphi' = \coalitione{A}{\omega,\varrho}\mnext \theta:$}{
      %\\ \quad 
     	%$\comp \ \M_{\varrho}$; 
     	$\sbrack{\varphi'}_{\M} = \ws(\Pre(A,\omega,\mapCap,\sbrack{\theta}_{\M_{\varrho}})  \cap  \dbrack{\varrho})$
    };
    \\
    %%% CASE 5 %%%%%
    \lCase{$\varphi' = \coalitione{A}{\omega,\varrho}\always \theta:$}{
       % \\ \quad 
        %$\comp \ \M_{\varrho};$
        $\rho=\sbrack{\true}_{\M_{\varrho}}; \tau=\sbrack{\theta}_{\M_{\varrho}}$ \; 
         \quad \!\!\!
         \lWhile{$\rho \not\subseteq \tau$}{$\rho=\tau;\; \tau = \Pre(A,\omega,\mapCap,\rho) \cap \sbrack{\theta}_{\M_{\varrho}}$ 
         \od}         
         \\ \quad
         $\sbrack{\varphi'}_{\M}=\ws(\rho \cap \dbrack{\varrho})$ 
    };
    \\
    %%% CASE 6 %%%%%
    \lCase{$\varphi' = \coalitione{A}{\omega,\varrho} \theta_1 \until \theta_2:$}{
        %\\ \quad 
        %$\comp \ \M_{\varrho}; $
         $\rho=\sbrack{\false}_{\M_{\varrho}}; \tau=\sbrack{\theta_2}_{\M_{\varrho}}$ \;
         \quad\!\!\!
         \lWhile{$\tau \!\not\subseteq\! \rho$}{$\rho=\rho \!\union\! \tau;\; \tau = \Pre(A,\omega,\mapCap,\rho) \!\cap\! \sbrack{\theta_1}_{\M_{\varrho}}\!$
            \od} \\
            \quad
           $\sbrack{\varphi'}_{\M}=\ws(\rho \cap \dbrack{\varrho})$
    };\\
    %\ecase
}
\od \\
\Return $\sbrack{\varphi'}_{\M}$;
\caption{BDI-ATLES symbolic model checking.}
\label{algo:BDI-ATLES}
\end{algorithm}}

So, given a BDI-ATLES model $\M \!\! = \!\! \tup{\agents,Q,\P,\actions,\atld,\mapFunc,\transFunc,\mapCap}$ and a goal assignment $\varrho$, the \defterm{goal-extended model} is a model $\M_{\varrho} \!\! = \!\! \tup{\agents,Q_{\varrho},\P,\actions,\atld_{\varrho},\mapFunc_{\varrho},\transFunc_{\varrho},\mapCap}$, where:
\begin{itemize}%\separation{-0.05in}
  \item $Q_{\varrho} \subseteq Q \times \prod_{\agt \in \A_{\varrho}} 2^{\varrho[\agt]}$ is the set of extended states, now accounting for the possible goals of BDI agents.
  When $q_\varrho = \tup{q,g_1,\ldots,g_{\card{\A_\varrho}}} \in Q_{\varrho}$, where $q \in Q$ and $g_i \subseteq \varrho[\agt_i]$, is an extended state, we use $\ws(q_\varrho) = q$ and $\goal(\agt_i,q_\varrho) = g_i$ to project $\M$'s world state and $\agt_i$'s goals.
  To enforce belief-goal consistency we require no agent ever wants something already true: there are no $q_\varrho \in Q_{\varrho}$, $\agt \in \A_{\varrho}$, and formula $\gamma$ such that $\mapFunc(\ws(q_\varrho)) \models \gamma$ and $\gamma \in \goal(\agt,q_\varrho)$.

  \item $\mapFunc_{\varrho}(q_\varrho)=\mapFunc(\ws(q_\varrho))$, for all $q_\varrho \in Q_\varrho$, that is, state evaluation remains unchanged.
  
  \item $\atld_{\varrho}(\agt,q_\varrho) = \atld(\agt,\ws(q_\varrho))$, that is, physical executability remains unchanged.

  \item $\transFunc_{\varrho}(q_\varrho,\moveVector) =
  \tup{q',g_1',\ldots,g_{\card{\A_\varrho}}'}$, where  
  $q' =\transFunc(\ws(q_\varrho),\moveVector)$ and
  $g_i' = \goal(\agt_i,q_\varrho) \setminus 
 \set{\gamma \mid \gamma \in \goal(\agt_i,q_\varrho), \mapFunc(q') \models \gamma}$, is the transition function for the extended model.
\end{itemize}

Model $\M_\varrho$ is like $\M$ though suitably extended to account for agents' goals under the initial goal-assignment $\varrho$. 
Observe that the transition relation caters for persistence of goals as well as dropping of achieved goals. Indeed, the extended system will never evolve to an (extended) state in which some agent has some true fact as a goal. Hence, the transition relation is well-defined within $Q_\varrho$ states.
%%
% and therefore ensures the traces of the extended model account for the second goal consistency constraint we defined earlier.
%%
More interesting, the extended model keeps the original physical executability of actions and, as a result, it accommodates both rational and irrational paths. However, it is now possible to discriminate between them, as one can reason about applicable plans in each state.
Finally, it is not difficult to see that the extended model is, in general, exponentially larger than the original one with respect to the number of goals $\max_{\agt \in \A}(|\varrho[\agt]|)$ and agents $\card{\A_{\varrho}}$.

As standard, we denote the states satisfying a formula $\varphi$ by $\sbrack{\varphi}$.
When the model is not clear from the context, we use $\sbrack{\varphi}_\M$ to denote the states in $\M$ that satisfy the formula $\varphi$.
We extend $\ws(\cdot)$ projection function to sets of extended states in the straightforward sense, that is, $\ws(S) = \bigcup_{q \in S} \set{\ws(q)}$. Thus,  $\ws([\varphi]_{\M_{\varrho}})$ denotes the set of all world states in $\M$ that are part of an extended state in $\M_\varrho$ satisfying the formula $\varphi$.
%%
%\textbf{Additionally, $\llbracket \cdot \rrbracket$ returns the states $\llbracket \varrho \rrbracket$  which have agents' goals as per a goal assignment $\varrho$. [[SS:  cannot parse it; two notations for the same thing?? $\llbracket \cdot \rrbracket$ and  $\llbracket \varrho \rrbracket$? 
%}
Also, $\dbrack{\varrho}$ denotes the set of extended states where the agents' goals are as per goal assignment $\varrho$; formally, $\dbrack{\varrho} = \set{q \mid q \in Q_{\varrho}, \forall \agt \in \A_{\varrho}:\goal(\agt,q)=\varrho[\agt]}$.

%%%%%%%%%%%%%%%%%%%%%%%%% BDI-Eval function %%%%%%%%%%%%%%%%%%%%%%%%%%%%%%%%%
%\input{alg-BDI-Eval}

Figure \ref{algo:BDI-ATLES} shows the model checking algorithm for \BDIOURS. It  is based on the symbolic model checking algorithm for ATL~\cite{AluraHK:JACM02-ATL} and ATLES~\cite{Walther:TARK07-ATLES}. 
The first three cases are handled in the same way as in ATL(ES).
To check the \BDIOURS coalition formulae $\BDIATLESCOLFORM$, we extend the model as above (relative to the formula's goal assignment $\varrho$), and then check the plain ATL coalition formula $\ATLCOLFORM$ in such extended model.
Note that only the set of states having the goals as per the initial goal assignment are returned---all agents' initial goals are active in the first state of any rational trace.

%%%%%%%%%%%%%%%%%%%%%%%%% BDI-Eval function %%%%%%%%%%%%%%%%%%%%%%%%%%%%%%%%%
%\input{alg-BDI-Eval}

Unlike standard ATL model checking, we restrict the agents' action choices as per their capabilities.
This is achieved by modifying the usual pre-image function $\Pre(\cdot)$ to only take into account actions resultant from agents' applicable plans.
More concretely, $\Pre(A,\omega,\mapCap,\rho)$ is the set of (extended) states from where agents in coalition $A$ can jointly force the next (extended) state to be in set $\rho$ no matter how all other agents (i.e., agents in $\A\setminus A$) may act and provided all BDI-style agents (i.e., agents with capabilities defined under $\omega$ and $\mapCap$) behave as such.
Formally:
\[
\begin{array}{l}
\Pre(A,\omega,\mapCap,\rho) = %\\
	\set{q \mid 
		\forall i \in A, \exists a_{i} \in \atld^+_\varrho(i,q,\omega,\Theta), \\ 
\phantom{xxxxxxxxxxxxxxxxxx}	
		\forall j \in \A\!\setminus\! A, \forall a_j \in \atld^+_\varrho(j,q,\omega,\Theta) \! : \!
%%		 
% \phantom{xxxxxxxxxxxxxxxxxx}
	\transFunc_\varrho(q,\tup{a_1,\ldots,a_{\card{\A}}}) \! \in \! \rho
	},
\end{array}
\]
where auxiliary function $\atld^+_\varrho(\agt,q,\omega,\Theta)$ denotes the set of all actions that an agent $\agt$ may take in state $q$ under capabilities as per defined in $\omega$ and $\Theta$:
{%\small
\[
	\atld^+_\varrho(\agt,q,\omega,\Theta) =  
	\begin{cases}
	\atld_\varrho(\agt,q)		& \text{if $\agt \not\in \A_\omega$} 
	\\[2ex]
	\atld_\varrho(\agt,q) \cap~ %\\
	\Rational(\agt,q,
		\displaystyle\bigcup_{c \in \omega[\agt]} \mapCap(c)) 
								& \text{if $\agt \in \A_\omega$}
	\end{cases}
\]
}

An action belongs to set $\atld^+_\varrho(\agt,q,\omega,\Theta)$ if it is physically possible (i.e., it belongs to $\atld_\varrho(\agt,q)$), and BDI-rational whenever the agent in question is a BDI agent. 
To capture the latter constraint, set $\Rational(\agt,q,\Pi)$ is defined as the set of all rational actions for agent $\agt$ in (extended) state $q$ when the agent is equipped with the set of plans $\Pi$:
\[
\Rational(\agt,q,\Pi) \! = \!\!
	\begin{cases}
  		\set{a \mid \phi[a]\psi \! \in\! \applicable(\agt,q,\Pi)}
  		& \!\!\!\! \text{if }\applicable(\agt,q,\Pi) \not= \emptyset
  		\\
		\set{\nop}	& \!\!\!\! \text{otherwise}
 	\end{cases}
\]
where $\applicable(\agt,q,\Pi) \! =  \! \set{\phi[a]\psi \in \Pi \mid 
  		\mapFunc(q) \models \phi,\gamma \in \goal(\agt,q),\psi \models \gamma}$ is the set of all applicable plans in $\Pi$ at state $q$.
So, summarising, function $\Pre(\cdot,\cdot,\cdot,\cdot)$ is an extension of the standard ATL $\Pre(\cdot)$ function in which the agents that have goals and capabilities defined---the BDI agents---do act according to those goals and capabilities.

It is clear that the modified version of $\Pre(\cdot)$ function does not alter the complexity of the underlying ATL-based algorithm. In fact, the variation is similar to that used for model checking ATLES, except that the action filtering does not come from strategy terms, but from agent plans.  
This means that the algorithm runs in polynomial time w.r.t. the size of model $\M_{\varrho}$ (which is exponential w.r.t. the original model $\M$).

\begin{theorem}\label{theo:complexity_full}
Model checking a \BDIOURS formula $\coalitione{A}{\omega,\varrho}\varphi$ (against a model $\M$) can be done in exponential time on the number of agents  $\card{\A}$ and goals $\max_{a \in \A}(|\varrho[a]|)$.
\end{theorem}
 
% Of course this is because we have used a succinct representation of models, close to ATL models. 
% %%
% Should we have included agents' goals explicitly in models, as done with intentions in ATL+intentions (ATLI)~\cite{jamroga:Springer2005}, the model checking problem would retain the polynomial time complexity from ATL. The same would apply if one just generalized ATLES to explicitly require all rational-strategies be part of the model.

Of course, should we have included agents' goals explicitly in models (rather than using a succinct representation), as done with intentions in ATL+intentions (ATLI)~\cite{jamroga:Springer2005}, the model checking problem would retain ATL's polynomial complexity. The same would apply if one just generalized ATLES to explicitly require all rational-strategies be part of the model.
The fact is, however, that generating such rational strategies by hand (to include them in models) will be extremely involved, even for small problems.
In addition, our approach decouples agent's mental attitudes from the physical ATL-like model, and enables reasoning at the level of formulae without changing the model.

We shall note that the exponentiality may not show up in certain applications.
In many cases, for example, one is interested in just one BDI agent acting in an environment. In that case, only such agent will be ascribed goals and capabilities. Since it arises due to agents with goals, the exponential complexity would therefore only be on the number of goals for such agent.
Similarly, in situations where all agents have a single goal to achieve (e.g., to pick gold), the model checking would then be exponential on the number of BDI agents only.
%%
%%
% Lastly, if there is information available about the capabilities and the goals of the other agents, then it might be possible in some situations to treat the problem as a two agent game by treating other agents as the environment agent.  
%%
In the next section we shall provide one interpretation of goals for which the model checking problem remains polynomial.

%%%%%%%%%%%%%%%%%%%%%%%%%%%%%%%%%%%%%%%%%%%%%%%%%%%%
\section{\BDIOURS with Maintenance Goals}\label{sec:specialCases}\label{sec:goals-maintenance}
%%%%%%%%%%%%%%%%%%%%%%%%%%%%%%%%%%%%%%%%%%%%%%%%%%%%

\newcommand{\BDIOURSM}{\ensuremath{\BDIOURS^{M}}}
\newcommand{\BDIOURSP}{\ensuremath{\BDIOURS^{P}}}

So far, we have worked on the assumption that agents have a set of ``flat'' \emph{achievement} goals, goals that the agent needs to eventually bring about.
One can however consider alternative views of goals that could suit different domains. In particular, we have considered achievement goals with \emph{priorities} and repetitive/reactive \emph{maintenance} goals. 
In the first case, the framework can be easily generalized to one in which goals can be prioritized without an increase in complexity~\cite{YadavSardina:CoRR12b}.
%%
% In the first case, the framework can be easily generalized to one in which goals can be prioritized---agents always act towards achieving a goal with highest priority, if at all possible--- without an increase in complexity; see~\cite{YadavSardina:CoRR12b}.

% In the first case, an agent is assumed to have a priority relation among its goals and always act---by means of an applicable plan---towards achieving a goal with highest priority, if at all possible. Thus, a gardening agent might want to both pluck fruits and collect dirt, but the former is of higher priority. 
% %%
% It is not difficult to adapt \BDIOURS to a more general variant \BDIOURSP that captures the priortized execution of plans, and show that the exponential complexity for model checking still remains~\cite{our-appendix}.

A more promising case arises when goals are given a maintenance interpretation, that is, (safety) properties that ought to be preserved temporally. For example, a Mars robot has the goal to always maintain the fuel level above certain threshold.
We focus our attention on the so-called \emph{repetitive} or \emph{reactive} maintenance goals~\cite{DuffHT06,Dastani:AAMAS06}: goals that ought to be restored whenever ``violated.'' Should the fuel level drop below the threshold, the robot will act towards re-fueling.  
This type of goals contrast with \emph{proactive} maintenance goals~\cite{DuffHT06}, under which the agent is expected to proactively avoid situations that will violate the goal. The fact is, however, that almost all BDI platform---like \JACK, \JASON, and \JADEX---only deal with the reactive version, thus providing a middle ground between expressivity and tractability.

Technically, to accommodate maintenance goals within \BDIOURS, one only needs to do a small adaptation of the semantics of the logic so that goals are not dropped forever once satisfied, but ``re-appear'' when violated.
%%
% Thus, the Mar robot agent will adopt the goal for re-fuelling when required.
%%
We refer to this alternate version of our logic as \BDIOURSM.
%%\footnote{Note we do now allow achievement and maintenance goals to co-exist in the current setting.} 
%%
Of course, the model checking algorithm discussed above also needs to be slightly adapted to deal with the new goal semantics. 
Interestingly, one only needs to adapt the definition of a goal-extended model $M_{\varrho}$ by re-defining components $Q_{\varrho}$ and $\transFunc_{\varrho}(q_\varrho,\moveVector)$; see~\cite{YadavSardina:CoRR12b} for details. 
% {\small
%   \[
%   \begin{array}{l}
%   Q_{\varrho} \! = \! 
%   \set{
%   	\tup{q,g_1,\ldots,g_{\card{\A_\varrho}}} \! \mid 
% %     	\phantom{xxxxxxxxxxxx}
% 		q \in Q, g_i = \varrho[\agt_i] \!\setminus \! \set{\gamma \! \mid \mapFunc(q)\models \gamma}  }; \\[2ex]
% %%
% \transFunc_{\varrho}(q_\varrho,\moveVector)= q'_\varrho$ \emph{iff} $\transFunc(\ws(q_\varrho),\moveVector) = \ws(q'_\varrho).		
%   \end{array}
%   \]
% }

% %%
% \begin{itemize}
%   \item The set of extended states $Q_{\varrho}$ is defined as follows:
% {\small
%   \[
%   \begin{array}{l}
%   Q_{\varrho} = 
%   \set{
%   	\tup{q,g_1,\ldots,g_{\card{\A_\varrho}}} \mid 
% %     	\phantom{xxxxxxxxxxxx}
% 		q \in Q, g_i = \varrho[\agt_i] \setminus \set{\gamma \mid \mapFunc(q)\models \gamma}  };
%   \end{array}
%   \]
% }
% 
%   \item $\transFunc_{\varrho}(q_\varrho,\moveVector)= q'_\varrho$ \emph{iff} $\transFunc(\ws(q_\varrho),\moveVector) = \ws(q'_\varrho)$.
% \end{itemize}

% It is not difficult to see that the transition function for the extended model is indeed well-defined, as each state in the original model $\M$ is uniquely extended by adding the corresponding goals for each agent. More importantly, the set of extended states is not longer exponential wrt the original set of states and, therefore, we get the following result.

\begin{theorem}
Model checking in \BDIOURSM can be done in polynomial time (w.r.t. the model and the formula).
\end{theorem}

Hence, for (reactive) maintenance goals, we retain ATL(ES) polynomial complexity.\footnote{Note the complexity of model checking ATLES is known only for memoryless strategies~\cite{Walther:TARK07-ATLES}.} 
Of course, this bound is tight, as \BDIOURSM subsumes ATL (just take $\omega=\varphi=\emptyset$ in every coalition formula) and model checking ATL is PTIME-complete~\cite{AluraHK:JACM02-ATL}.

\section{Discussion}\label{sec:conclusions}

We have developed an ATL-like logic that relates closely to the BDI agent-oriented programming paradigm widely used to \emph{implement} multi-agent systems.
In the new logic, the user can express the capability of agents equipped with know-how knowledge in a natural way and can reason in the language about \emph{what agents can achieve under such capabilities}. 
Besides the general framework with standard achievement goals, we argued that one could instead appeal to goals with priorities or a special type of maintenance goals. We provided algorithms for model checking in such a framework and proved its (upper-bound) complexity in the various cases.
Overall, we believe that this work is a first principled step to bring together two different fields in the area of multi-agent systems, namely, verification of strategic behaviour and agent programming.

% Although our framework is abstract and hence not linked to any \emph{specific} BDI programming framework, it is possible to prove  its relationship with most such languages under plausible assumptions (e.g., use of conjunctive goals only, linear execution model, etc.) 

%Though not included for lack of space, a graphical representation of the technical notions used and a reduction of our logic to a version of ATLES with nondeterministic commitment functions can be found in~\cite{us-aaai11}. We shall include this content in the final version with an extra \nolinebreak page.

The framework presented here made a number of assumptions requiring further work.
Due to valuation function $\V$ in a structure, all agents are assumed to have full shared observability of the environment. This is, of course, a strong assumption in many settings.
We considered here basically reactive plans, akin to the language of \GOALBDI~\cite{Hindriks:Goal07}, certain classes of 2APL/3APL~\cite{Dastani:JAAMAS08-2APL,Hindriks:JAAMAS99-3APL}, reactive modules~\cite{Baral:ETAI98}, and universal plans~\cite{Schoppers:IJCAI87-UniversalPlans}. We would like to explore the impact of allowing plan bodies having sequences of actions, and more importantly, sub-goaling, as well as the possibility of agents imposing (new) goals to other agents, via so-called BDI \emph{messages}.
Also, in the context of complex plan bodies, one could then consider both a linear as well as interleaved execution styles of plans within each agent (for its various goals).
Most of these issue appear to be orthogonal to each other, and hence can be investigated one by one. With the core framework laid down, our next efforts shal focus on the above issues, as well as proving whether the complexity result provided in Theorem~\ref{theo:complexity_full} is tight.

We close by noting that, besides ATLES, our work has strong similarities and motivations to those on \emph{plausibility} \cite{jamroga:AAMAS07} and \emph{intention}~\cite{jamroga:Springer2005} reasoning  in ATL. Like ATLES, however, those works are still not linked to any approach for the actual development of agents, which is the main motivation behind our work. Nonetheless, we would like to investigate how to integrate plausibility reasoning in our logic, as  it seems orthogonal to that of rational BDI-style behavior. Indeed, the plausibility approach allows us to focus the reasoning to certain parts of an ATL structure using more declarative specifications.

% 
% ks done earlier on plausibility, with plausibility in this setting be defined by the agent's capabilities. But, the frameworks that define \emph{plausibility}~\cite{jamroga:AAMAS07} and  require the strategies to be defined before hand. 

\small

% \paragraph{Acknowledgments} 
% We thank Lawrence Cavedon for earlier discussions on the topic of this paper. We acknowledge the support of the Australian Research Council under grant DP120100332.
% \vspace*{-.8cm}
% \bibliographystyle{plain}
% \bibliographystyle{abbrvnat}	%% First name abbreviated G. De Giacomo
% \bibliography{bib-ssardina}

\normalsize
\newpage
\appendix
\section{Rational strategies}\label{sec:rational_traces}

Given a plan library $\Pi$ and an initial goal base $\G$ for an agent $\agt$ in structure $\M$, we are to characterize those strategies for $\agt$ within $\M$ that represent rational behaviors: \emph{the agent tries plans from $\Pi$ in order to bring about its goals $\G$ given its beliefs}~\cite{Bratman:CI88,Rao:KR92}.

While technically involved, the idea to define set $\Sigma^\agt_{\Pi,\G}$ is simple: first identify those paths in $\M$ that display rational behavior for the agent; second consider rational strategies those that will always yield rational paths.
We do this in three steps. First, we identify constraints on how the goals of an agent can evolve in a path. Second, we define what it means for a plan to be tried by an agent in a path. Third, we identify those (rational) paths that result from an agent's deliberation process. 
%%
% We close the loop by defining plan compatible strategies as those which result in such rational paths.

Before we start, we extend the notion of paths to account for the actions performed. A \defterm{trace} is a finite sequence of alternating states and actions $\lambda^+=q_0\act_1 q_1\cdots\act_\ell q_\ell$ such that $q_0q_q\cdots q_\ell$ is a (finite) path in $\M$.
As with paths, we use $\lambda^+[i]$ and $\lambda^{+}\tup{i}$ to denote the $i$-th state $q_i$ and the $i$-th action $\act_i$, respectively, in trace $\lambda^+$. The length $\card{\path^+}$ of a trace is the number of actions on it; hence it matches the length of the underlying path.

\begin{example}
Figure~\ref{fig:traces} depicts two possible traces $\path^+_1$ and $\path^+_2$ for agent $\Ag$ that are compatible with strategies $f^1_{\Ag}$ and $f^2_{\Ag}$, respectively.
The agent has $\WAg$ as its initial goal, and is equipped with capabilities $\Navigation$, $\Pick$ and $\Drop$.
Trace $\path^+_1$ is resultant from the agent executing actions as per its applicable plans, as evident from the plan labeling.
For example, at the state $q_1$, the agent is collocated with the gold, and so executes the \aPick action as per plan $\agC\wedge\gC[\aPick]\gB$.
Consequently, strategy $f^1_{\Ag}$ is rational as it yields rational trace $\path^+_1$.
On the other hand, trace $\path^+_2$ does not obey the rationality constraints: the agent executes the $\nop$ action at $q_0$, although there is in fact an applicable plan available.

\end{example}

% We now have all the technical machinery to define the set of rational traces and the set $\Sigma^{\agt}_{\Pi,\G}$ of rational  strategies used to define the semantics of \BDIOURS in the previous section.
%%
	
%%
Technically, $\path^+$ is a \defterm{rational trace} for an agent $\agt \in \A$ equipped with a plan-library $\Pi$ and having an initial goal base $\G$, if for all $i<|\path^+|$, either:
\begin{itemize}
  \item there is an applicable plan $\plan{} \in \Pi$ (for some goal $\gamma \in \goalFunc_\G(\path^+,i)$) such that $i \in \exec_\agt(\plan{},g_\G,\lambda^+)$; or
  
  \item there is no applicable plan relative to the goal-marking $\goalFunc_\G$ and $\path^+\tup{i+1} = \nop$.
\end{itemize}

We denote $\zeta^{\agt}_{\Pi,\G}$ the set of all rational traces for agent $\agt$ with library $\Pi$ and goal base $\G$. 
Also, to link traces and strategies, we define $\tr(\path,f_\agt)$ to be the partial function that returns the trace induced by a path $\lambda$ and a strategy $f_\agt$, if any.
Formally, $\tr(\path,f_{\agt})= q_0\act_1 q_1\ldots\act_{|\path|} q_{|\path|}$ iff for all $k \leq |\path|$: \emph{(i)} $q_k=\path[k]$; and \emph{(ii)} there exist a joint-move $\moveVector \in \D(q_k)$ such that $\transFunc(q_k,\moveVector)=q_{k+1}$ and $f_{\agt}(\path[0,k]) = \moveVector[\agt] = \act_{k+1}$ (where $\moveVector[\agt]$ denotes agent $\agt$'s move in joint-move $\moveVector$).

Finally, the set of \defterm{rational strategies} is defined as follows:
\[
\begin{array}{l}
\Sigma^{\agt}_{\Pi,\G} = 
	\set{f_\agt \mid 	\bigcup_{\path \in \overline{\Lambda}}\tr(\path, f_\agt) \subseteq \zeta^{\agt}_{\Pi,\G} }, 
\end{array}
\]

\noindent
where $\overline{\Lambda} \subseteq \Lambda$ is the set of finite paths in $\M$.
That is, a rational strategy $f_\agt$ is one that only yields rational traces.

% Formally, $\tr(\path,f_{\agt})= q_0\act_1 q_1\ldots\act_{|\path|} q_{|\path|}$ iff for all $k \leq |\path|$:
% %%
% \begin{itemize}
%   \item $q_k=\path[k]$;
%   
%   \item there is a joint-move $\moveVector \in \D(q_k)$ such that $\transFunc(q_k,\moveVector)=q_{k+1}$ and $f_{\agt}(\path[0,k]) = \moveVector[\agt] = \act_{k+1}$.
% \end{itemize}
%\vspace{0.2cm}

\subsubsection{Goal evolution in traces}\label{sec:goal-evolution}

BDI agents achieve their goals by means of acting as per their plans. 
In this section, we shall identify what the possible \emph{achievement} goals~\cite{Dastani:AAMAS06} an agent may have at each moment in a path, under a blind commitment strategy~\cite{Rao:KR91}---the most common strategy in BDI programming platforms---in which a rationality principle states that an agent drops the goals it has already achieved.
We discuss \emph{maintenance} goals in the next section as a special case.

To that end, we make use of so-called \defterm{goal-marking} function $g_\G(\path^+,i)$, for an agent with an initial goal base $\G$, which outcomes  the ``active" goal base of the agent at moment $\path^+[i]$ in trace $\path^+$. 
As expected, the goal marking function returns the unfulfilled subset of initial goals.
Formally,
\[
\begin{array}{l}
g_\G(\path^+,i) = 
	\set{\gamma \mid \gamma \in \G,\; (\neg\exists j \leq i) \ \mapFunc(\path^+[j])\models \gamma}.
\end{array}
\]

%%%%%%%%%%%%%%%%%%%%% OLD GOAL MARKING CONSTRAINTS %%%%%%%%%%%%%%%%%%
%%
%As expected, a goal-marking function must obey some basic rationality constraints in terms of goal persistence: 
%%
%\begin{itemize}
  
%  	\item  $g_\G(\path^+,0)= \G$, that is, the agent's initial goal base is $\G$;
  
%	\item  for all $i \leq |\path^+|$ and $\gamma \in g_\G(\path^+,i)$, $\mapFunc(\path^+[i]) \not\models \gamma$, that is, the agent does not have (already) achieved goals.

%    \item for all $i \leq |\path^+|$  we have that:
%    \begin{itemize}
%      \item $g_\G(\path^+,i) \subseteq \G$, that is, every goal comes from the agent's initial goal base (no new goals adopted); and

%      \item for all $\gamma \in g_\G(\path^+,i)$ and all $j\leq i$, $\mapFunc(\path^+[j]) \not\models \gamma$, that is, the agent does not have goals that have already been achieved;
       %%
      %%
%      \item for all $\gamma \in \G \setminus g_\G(\path^+,i)$, there  exists $j \leq i$ such that $\mapFunc(\path^+[j]) \models \gamma$, that is, the agent drops only goals that have been achieved.
%    \end{itemize}
%\end{itemize}
%%%%%%%%%%%%%%%%%%%%%%%%%%%%%%%%%%%%%%%%%%%%%%%%%%%%%%%%%%%%%%%%%%%%%%%%%

%Observe that these two constraints only capture half of the single-minded notion of commitment. Indeed, they do not detail on how an agent may abandon goals besides when achieved or how it may adopt new (sub)goals.  
%%
%To complete the picture, we need to take plans into consideration.

%Observe that we have not defined any agent \emph{commitment} constraints. 
%
In this setting, the agent picks an active goal and executes a plan for it. If the plan fails to achieve the goal, the goal remains active and the agent can either pursue the same goal or a different one. On the other hand, if the plan succeeds in bringing about the goal, then the achieved goal is removed from the agent's active goal set.

% Note that, the path $\lambda_l$ should result from the agent $\agt_i$ choosing actions as per the strategy $f_{\agt_i}$ from the state $q_0$. If the path $\lambda_l$ is not compatible with $f_{\agt_i}$, that is $\exists j : \forall \moveVector \in \D(q_j) \textit{ where } \moveVector[i]=f_{\agt_i}(\path_l[0,j]) \ \textit{, } \ \transFunc(q_j,\moveVector)\neq q_{j+1}$, then the function $tr(\path_l,f_{\agt_i})$  returns $\epsilon$.

\subsubsection{Plan executions in traces}\label{subsec:planrecog}

Agents developed under the BDI paradigm are meant to execute actions as per the plans/know-how available to them. We shall next define what it means for a trace to include an execution of a plan. 
%%

% More precisely, given a  plan-rule $\phi[\alpha]\psi$, an agent may decide to execute plan $\alpha$ if bringing about $\psi$ will achieve some of its current goals.

When it comes to selecting plans for execution, there are generally two core notions in BDI programming.
A plan is relevant if its intended effects are enough to bring about some goal of the agent. Technically, given a trace $\path^+$ and the goal-marking function $g$, we say that a plan-rule $\plan{}$ is \defterm{relevant} at moment $\path^+[i]$ in the trace, with $0 \leq i \leq \card{\path^+}$, if there exists $\gamma \in g(\path^+,i)$ such that $\psi \models \gamma$.
Furthermore, the plan is \defterm{applicable} at $\path^+[i]$ if it is relevant and its context condition holds true, that is, $\mapFunc(\path^+[i])\models \phi$.

So, we identify the moments in which a particular plan may have been executed in a trace by an agent.
%%
% The function $\exec_\agt(\plan{},g,\lambda^+)$ denotes the locations in trace $\lambda^+$ that stand for an execution of  plan  by agent . 
%%
Formally,  $\exec_\agt(\plan{},g,\lambda^+)$ is the set of indices $i$ such that \emph{(i)} $\lambda^+\tup{i+1} = \alpha$; and \emph{(ii)} $\plan{}$ is applicable at $\lambda^+[i]$ under goal-marking $\goalFunc$.

\subsubsection{Rational traces and rational strategies}

We now have all the technical machinery to define the set of rational traces---those that can be explained by the agent acting as per its available plans in order to achieve its goals relative to its beliefs---and the set $\Sigma^{\agt}_{\Pi,\G}$ of rational  strategies used to define the semantics of \BDIOURS.

Initially, an agent has a set of goals (initial goal base) that she wants to bring about. The agent then chooses one goal to work on, and selects an applicable plan for such goal from its plan library for execution.
If the plan successfully brings about the goal, then the agent deems the goal achieved and the plan finished.
%%
%On the other hand, if the plan fails to achieve the goal, then the agent executes another applicable plan for the \emph{same} goal (even the same failed plan if still applicable), thus realizing its commitment to the goal.  
%%
%When the agent has no applicable plan for a (sub)goal, the agent deems the (sub)goal impossible and drops it. 
%
Traces that can be ``explained'' in this way are referred to as \emph{rational traces}.

%%%%%%%%%%%%%%%%%%% EXAMPLE %%%%%%%%%%%%%%%%%%%%

%\begin{figure}[!t]
%\begin{center}
%\resizebox{\columnwidth}{!}{\input{fig-traces}}
%\end{center}
%\caption{Traces $\path^+_1$ and $\path^+_2$ for an agent $\Ag$ resultant from strategies $f^1_{\Ag}$ and $f^2_{\Ag}$ respectively. 
%
%The states are labeled as per the model in figure~\ref{fig:goldModel}. 
%
%The strategy $f^1_{\Ag}$ is rational as it yields rational trace $\path^+_1$.}
%\label{fig:traces}
%\end{figure}

%%%%%%%%%%%%%%%%%%%%%%%%%%%%%%%%%%%%%%%%%%%%%%%%%%%%
\section{Priorities and Maintenance Goals }\label{sec:specialCases}
%%%%%%%%%%%%%%%%%%%%%%%%%%%%%%%%%%%%%%%%%%%%%%%%%%%%

So far, we have studied \BDIOURS under the assumption that agents have a set of ``flat'' \emph{achievement} goals, goals that the agent needs to eventually bring about.
In this section, we consider our logic under two alternative views of the agent's goal base that could suit different domains, namely, achievement goals with \emph{priorities} and repetitive/reactive \emph{maintenance} goals. Interestingly, the model checking problem for the latter special case remains polynomial.

\subsection{Goals with priority}\label{sec:goals-priority}

The \BDIOURS framework defined above treats all goals with equal importance, in that a BDI agent may choose to execute an applicable plan for any active goal.
In many situations, though, an agent may prefer achieving certain goals first.
Thus, a gardening agent might want to both pluck fruits and collect dirt, but the former is of higher priority. 
Similarly, a sales agent will generally prefer attending a customer or a phone call to do some back-office task.

To accommodate goal bases with priorities,  we extend the notion of goal assignment ($\varrho$) as tuples of the form  $\tup{\agt:(\Gamma,\prec)}$, where $\agt \in \agents$, $\Gamma$ is a finite set of boolean formulas over $\P$, and $\prec$ is a partial-order relation over  $\Gamma$ specifying the priority over the goals.
As expected, an initial goal base $\G = (\Gamma_0,\prec_0)$ consists of a set of goals formulas $\Gamma_0$ and a partial-order priority relation $\prec_0$ over such set.
We use $\prec_{\agt}$ to denote the goal priority of $\agt \in \A_{\varrho}$ in goal assignment~$\varrho$.

Roughly speaking, an agent always acts---by means of an applicable plan---towards achieving a goal with highest priority.
Should that be impossible (i.e., there is no applicable plan for such goals), the agent suspends those goals and tries to act towards a next preferred goal, and so on.
Note that, being achievement goals, these are meant to be dropped whenever achieved. Moreover, goals at any level of importance could be achieved as (un-intended) side effects of the agent's own actions or even other agents' actions.

To achieve all this, we update the definition of rational traces and strategies as follows. 
A trace $\path^+$ is \emph{rational} for an agent $\agt \in \A$ having an initial goal base $\G=(\Gamma_0,\prec_0)$ and a plan library $\Pi$, if for all $i<\card{\path^+}$ either: (here, $g_{\G}(\cdot,\cdot)$ is the corresponding goal-marking function)
\begin{itemize}\separation{-0.05in}
  \item there is an applicable plan $\plan{} \in \Pi$ at state $\path^+[i]$ for a goal $\gamma \in \goalFunc_\G(\path^+,i)$ such that $i \in \exec_\agt(\plan{},g_\G,\lambda^+)$, and there is no $\gamma' \in g_{\G}(\path^+,i)$ with an applicable plan at $\path^+[i]$ and such that $\gamma' \prec_0 \gamma$; or
  
  \item there is no applicable plan at $\path^+[i]$ for any goal in $\goalFunc_\G(\path^+,i)$ and $\path^+\tup{i+1} = \nop$.
\end{itemize}

Observe that the only difference with the original notion of rational traces is that an agent cannot act on a lower priority goal if it has an applicable plan for a higher priority goal.
All other notions, including goal-marking functions and plan execution in traces remain the same. 
We refer to this new variant as \BDIOURSP.

Now, a model checking algorithm for \BDIOURSP can be obtained by slightly modifying the $\Pre(\cdot,\cdot,\cdot,\cdot)$ function in the algorithm for \BDIOURS.
Instead of considering all applicable plans, the $\Pre$ function needs to just take into account applicable plans for the most preferred goal(s).
This can be easily achieved by using function $\applicable^+(\agt,q,\Pi)$ in the definition of  $\Rational(\agt,q,\Pi)$: 
%%
%\[
%\begin{array}{l}
%\applicable^+(\agt,q,\Pi) = 
%	\set{\planf{\phi}{a}{\psi} \in \applicable(\agt,q,\Pi) \mid \\
%	\phantom{xxxxxxxxxxxx}
%  		(\neg \exists \planf{\phi'}{a'}{\psi'} \in \applicable(\agt,q,\Pi)).\psi' \prec_{\agt} \psi}.
%\end{array}
%\]
\[
\begin{array}{l}
\applicable^+(\agt,q,\Pi) = \\
    \phantom{xx}
    \set{\planf{\phi}{a}{\psi} \in \applicable(\agt,q,\Pi) \mid \\
    \phantom{xxxxx}
     (\exists \gamma \in \goal(\agt,q)).\psi \models \gamma, \\
    \phantom{xxxxxxx} 
     (\neg \exists \planf{\phi'}{a'}{\psi'} \in \applicable(\agt,q,\Pi),\gamma'\in \goal(\agt,q)) \\
    \phantom{xxxxxxxxxxxx}
        \gamma' \prec_\agt \gamma \wedge \psi' \models \gamma'  
     }.
\end{array}
\]

That is, $\applicable^+(\agt,q,\Pi)$ denotes the set of all \emph{highest-priority} applicable plans at $q$: there is no plan 
$\planf{\phi'}{a'}{\psi'}$ applicable at $q$ and serving a higher-priority goal $\gamma'$.

%%
%\textbf{IS THIS OK? the prioerities will not specified in terms of formulas in plans. Something more involved may be needed here} 

% Formally, $\Pre(A,\omega,\mapCap,\rho)$ contains state $q$ if for every player $a \in A$ there exists a move  $j_a \in \atld_{\varrho}(a,q)$ such that for all players $b \in \A \setminus A$ and moves $j_b \in \atld_{\varrho}(b,q)$ we have that $\transFunc_{\varrho}(q,\tup{j_1,\ldots,j_{|\A|}}) \in Q$ where for all players $c \in \A_{\omega}$, $\exists \phi[j_c]\psi \in \cup_{m \in \omega[c]}\mapCap(m)$ such that the plan $\phi[j_c]\psi$ is applicable at $q$ and there does not exist an applicable plan $\phi'[\alpha']\psi' \in \cup_{m \in \omega[c]}\mapCap(m)$ where $\psi' \succ \psi$ and $\alpha' \in \atld_{\varrho}(c,q)$, or $j_c=\nop$ if there is no applicable plan at $q$.

We note that the complexity of model checking this variant with goal priorities remains the same as that of \BDIOURS. Indeed, the size of the extended model is still exponential w.r.t. the original model. This is because the agent might achieve lower priority goals as non-intended side effects, and as a result, one still needs to extend the original states with the powerset of the initial goal base.
%%
% \begin{proposition}
% Model checking in \BDIOURSP is exponential on the number of agents and agents' goals.
% \end{proposition}

Nonetheless, the \BDIOURSP variant is more general; if agents have equal priorities on all their goals, then it is equivalent to the original \BDIOURS framework. 
On the other hand, priorities on goals can yield very different abilities among agents and coalitions. In fact, the ability of an agent (or coalition) may depend on the priority of other agents: an gold miner agent may be able to pick gold pieces \emph{only if} the opponent prefers exploring grid to picking gold. 

Lastly, it is not difficult to provide different semantics to goal priorities. One can imagine agents dropping goals or blindly waiting for a plan to become applicable when none is available, rather than suspending the goal and acting on lower priority goals.
While it is straightforward to capture such semantics, the exponential complexity for model checking still remains.

\subsection{Maintenance goals}\label{sec:goals-maintenance}

In order to force goals to ``re-appear'' when violated, we re-define the goal-marking function $g_\G(\cdot,\cdot)$, which determines the goal base of an agent at a given $i$-th moment in a trace $\path^+$, as follows:
{
\[
\begin{array}{l}
	g_\G(\path^+,i) = \set{\gamma \mid \gamma \in \G,\;  \mapFunc(\path^+[i])\models \gamma}.
\end{array}
\]
}
Importantly, the rest of the definitions for plan execution and rational strategies remain exactly the same.
We shall refer to this alternate version of our logic as \BDIOURSM.\footnote{Note we do now allow achievement and maintenance goals to co-exist in the current setting.} 

In terms of the model checking algorithm, the only change required is in the definition of a goal-extended model $M_{\varrho}$, which is defined as above except for the following:
\begin{itemize}
  \item The set of extended states $Q_{\varrho}$ is defined as follows:
  \small
  \[
  \begin{array}{l}
  Q_{\varrho} = \\
    	\phantom{x}
  \set{
  	\tup{q,g_1,\ldots,g_{\card{\A_\varrho}}} \mid \\
    	\phantom{xxxxxxxxxxxx}
		q \in Q, g_i = \varrho[\agt_i] \setminus \set{\gamma \mid \mapFunc(q)\models \gamma}  };
  \end{array}
  \]
	
  \item $\transFunc_{\varrho}(q_\varrho,\moveVector)= q'_\varrho$ \emph{iff} $\transFunc(\ws(q_\varrho),\moveVector) = \ws(q'_\varrho)$.
\end{itemize}

It is not difficult to see now that the set of extended states is no longer exponential w.r.t. the original set of states.

\end{document}